\documentclass[journal]{IEEEtran}
\ifCLASSINFOpdf
\else
\fi

\usepackage{amssymb}
\usepackage{booktabs}
\usepackage{times}
\usepackage{epsfig}
\usepackage{graphicx}
\usepackage{amsmath}
\usepackage{amssymb}
\usepackage{multirow}
\usepackage{booktabs}
\usepackage{pifont}
\usepackage{comment}
\usepackage{mathtools}
\usepackage{url}
\usepackage{bm}
\usepackage{slashbox}
\usepackage{listings}

\newcommand{\cmark}{\ding{51}}%
\begin{document}
%
\title{Joint Forecasting 
  of Features and Feature Motion\\
  for Dense Semantic Future Prediction\thanks{This 
    work has been funded by Rimac Automobili.
    This work has also been supported 
    by the Croatian Science Foundation
    under the grant ADEPT and
    European Regional Development Fund 
    under the grant KK.01.1.1.01.0009 DATACROSS.}}
\title{Dense Semantic Forecasting in Video \\
by Joint Regression 
of Features and Feature Motion
\thanks{This 
    work has been funded by Rimac Automobili.
    This work has also been supported 
    by the Croatian Science Foundation
    under the grant ADEPT and
    European Regional Development Fund 
    under the grant KK.01.1.1.01.0009 DATACROSS.}}

%
%
%
\author{
  Josip Šarić, \and
  Sacha Vražić, \and
  Siniša Šegvić %
  \thanks{
  J.\ Šarić and S.\ Šegvić are with University of Zagreb, Faculty of Electrical Engineering and Computing (e-mail: \{josip.saric, sinisa.segvic\}@fer.hr).}%
  \thanks{
  S. Vražić is with Rimac Automobili, Sveta Nedelja, Croatia (e-mail: sacha.vrazic@rimac-automobili.com).}
}

%
%

\markboth{IEEE Transactions on Neural Networks and Learning Systems}%
{Šarić \MakeLowercase{\textit{et al.}}: Bare Demo of IEEEtran.cls for IEEE Journals}
%



\maketitle


\begin{abstract}
Dense semantic forecasting 
anticipates future events 
in video by inferring pixel-level semantics 
of an unobserved future image. 
We present a novel approach 
that is applicable to various 
single-frame architectures and tasks. 
Our approach consists of two modules.
Feature-to-motion (F2M) module
forecasts a dense deformation field
that warps past features into their future positions.
Feature-to-feature (F2F) module
regresses the future features directly
and is therefore able to account for emergent scenery.
The compound F2MF model
decouples the effects of motion
from the effects of novelty
in a task-agnostic manner.
We aim to apply F2MF forecasting
to the most subsampled and
the most abstract representation
of a desired single-frame model.
Our design takes advantage
of deformable convolutions and
spatial correlation coefficients
across neighbouring time instants.
We perform experiments
on three dense prediction tasks:
semantic segmentation,
instance-level segmentation,
and panoptic segmentation.
The results reveal state-of-the-art 
forecasting accuracy
across three dense prediction tasks.
\end{abstract}


%
\IEEEpeerreviewmaketitle

\section{Introduction}
Visual perception is an important challenge
towards development of 
autonomous robots and vehicles.
Recent discovery of deep learning methods
triggered a great progress in single-image 
dense prediction tasks
such as instance segmentation \cite{he2017mask}
or panoptic segmentation \cite{Cheng_2020_CVPR}.
However, the reasoning of an intelligent agent 
must not be limited 
to the present moment in time
since consequences of our current actions
and the goals of our missions 
occur in the future.
Consequently, anticipation of future events
could be an important ingredient 
towards making our present systems
better and more intelligent.

\begin{figure}[t]
  \centering
  \includegraphics[width=\columnwidth]
    {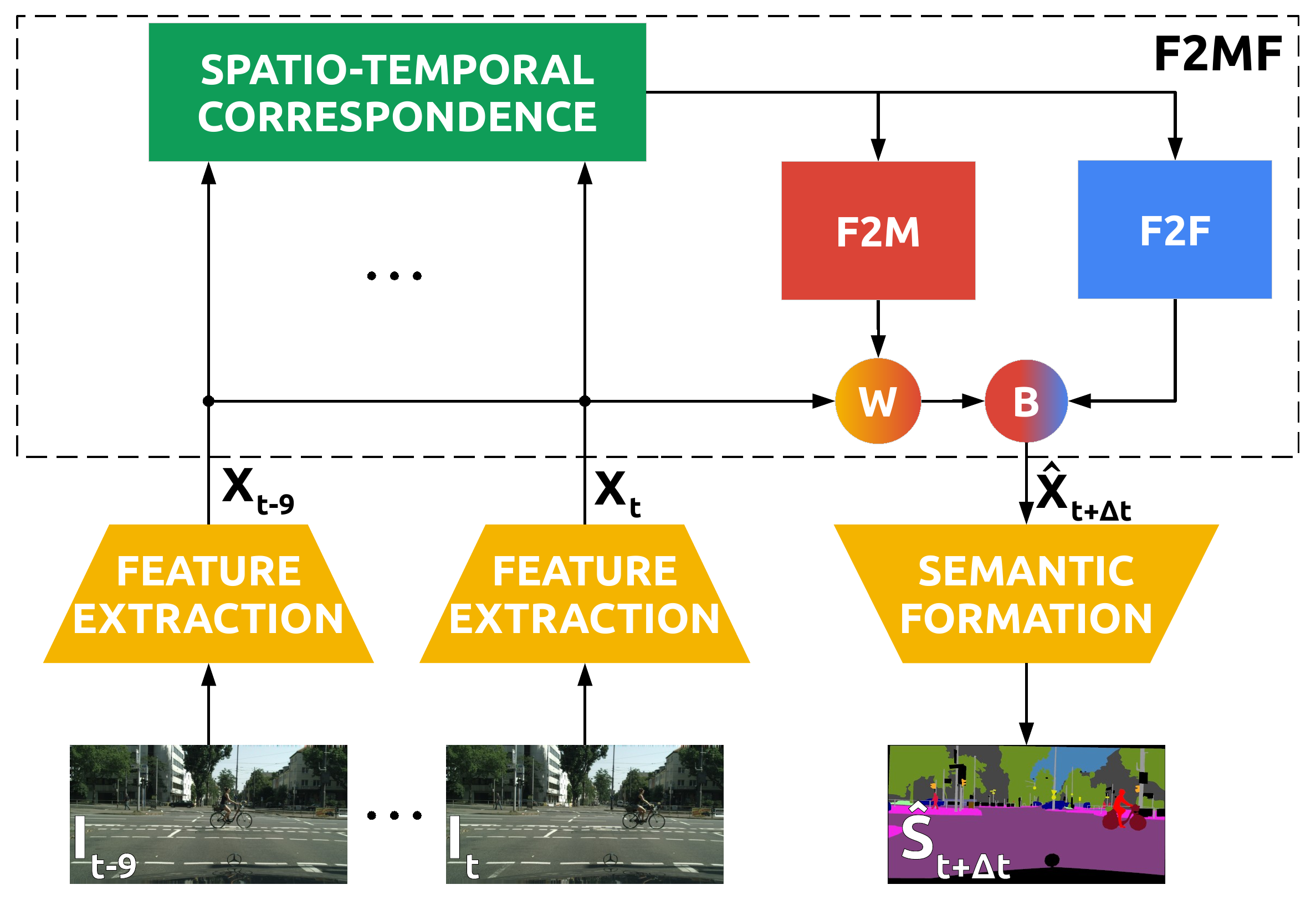}
  \caption{Overview of the proposed 
    F2MF forecasting.
    Low-resolution features 
    $\mathbf{X}_\tau$
    are extracted from
    observed RGB images 
    $\mathrm{I}_\tau$,
    $\tau\in\{t-9,t-6,t-3,t\}$
    by a pre-trained recognition module.  
    The features are enriched
    with their spatio-temporal correlations
    and forwarded to F2M and F2F modules 
    which specialize for forecasting 
    previously observed and novel scenery.
    The forecasted future features
    $\mathbf{\hat{X}}_{t+\Delta t}$
    are a blend (B) of F2M and F2F outputs.
    Dense semantic predictions
    $\mathbf{\hat{S}}_{t+\Delta t}$ 
    are finally formed throughout 
    a pre-trained upsampling module.
  }
  \label{fig:intro}
\end{figure}
Early visual forecasting methods
relied on handcrafted models 
of scene dynamics \cite{davison07pami}.
However, this approach is prone 
to systematic errors due to
insufficient modeling accuracy.
For instance, popular 
camera models \cite{zhang16mva}
are unable to express arbitrary lens distortions \cite{krevso2015improving}.
Likewise, object-level forecasting 
by geometric filtering \cite{reuter14tsp} 
is vulnerable to propagation 
of detection and association errors.
Consequently, recent work attempts 
to implicitly capture 
the laws of scene dynamics
through deep learning in video
\cite{oprea2020pami}.
This can be carried out by 
forecasting either future RGB frames \cite{mathieu16iclr,gao2019disentangling},
or the corresponding semantic content
\cite{alahi2016social,luc2017predicting,luc2018predicting}.

Semantic forecasting 
is an appealing approach
since decision-making systems 
mainly care about the \emph{content}
of future scenes
rather than about their \emph{appearance}
\cite{yao19icra}.
Empirical evidence indicates 
that direct semantic foreasting 
performs much better than 
inferring from a forecasted RGB frame,
while requiring significantly 
less computational effort \cite{luc2017predicting}.  
Intuition suggests 
similar conclusions. 
It seems appropriate 
to first figure out 
what is going to happen 
at an abstract level
before moving on to RGB pixels 
and the associated details of lighting, 
reflectance and surface normals.
Indeed, it appears that
semantic forecasting 
should be a prerequisite 
for RGB forecasting 
rather than vice versa.

There are multiple approaches 
for expressing semantic forecasting.
Semantics-to-semantics
forecasting (S2S)
maps observed semantic maps
into their future counterparts
\cite{luc2017predicting}.
Motion-to-motion forecasting (M2M)
receives the optical flow 
between observed images and produces
the optical flow between the future image
and the last observed image
\cite{terwilliger2019recurrent}.
Finally, feature-to-feature forecasting (F2F) operates
on abstract convolutional features 
\cite{luc2018predicting}.
There are several advantages of F2F forecasting 
with respect to the remaining two approaches.
In comparison with S2S,
it offers better spatio-temporal correspondence, 
more expressive features,
and task-agno\-sti\-cism.
In comparison with M2M,
it offers semantic reasoning 
and in-painting the novel scenery.

This paper builds on 
previous 
feature-level forecasting
approaches \cite{luc2018predicting,sun2019predicting}
and enriches them with 
the following contributions.
We propose 
F2M forecasting (feature-to-motion)
which expresses future features 
by warping their current counterparts 
throughout a dense deformation field. 
This formulation expresses feature-level forecasting 
as a causal relationship 
between the past and the future. 
However, F2M forecasting is unable 
to predict emergence of novel scenery. 
Hence, we finally propose 
F2MF forecasting (feature-to-motion-and-feature) 
by blending F2M and F2F 
with densely regressed weight factors
as illustrated in Figure \ref{fig:intro}.
This allows our method to foresee
emergence of unobserved scenery,
and to exploit that information
for decoupling variation due to novelty
from variation due to motion.
We account for geometric nature of our task
by leveraging spatio-temporal 
correlation coefficients
and deformable convolutions.
We show that dense semantic forecasting
can be expressed in terms
of the most compressed representation
of the single-frame model, and that such organization 
can lead to competitive accuracy.
The resulting generalization performance 
surpasses the state-of-the-art 
by a wide margin.
Our method is very-well suited
for real-time implementation
due to low computational overhead 
with respect to single-frame prediction
\cite{vsaric2019single}.
The method can be adapted
for other dense prediction tasks,
which we explain next.

In addition to contributions 
from previous conference accounts
\cite{vsaric2019single,Saric_2020_CVPR},
here we extend F2MF forecasting 
with two new single-frame architectures
\cite{he2017mask,Cheng_2020_CVPR},
and show that single-level forecasting
can preserve instance-related information.
Thus, this is 
the first forecasting method 
that can be applied 
to three dense prediction tasks: 
semantic segmentation,
instance-level segmentation, 
and panoptic segmentation. 
We note that transition 
towards instance-aware tasks 
is not trivial since 
the corresponding metrics (AP and PQ) 
aggregate instance-level 
instead of pixel-level recognition quality.
Hence, correct recognition of small objects 
becomes much more important 
than in semantic segmentation.
This paper also presents computational advantages
of our method by comparing its complexity
with other dense semantic forecasting approaches.
Finally, it demonstrates that our method 
is able to generalize over different 
cameras, resolutions and framerates,
as well as to produce longer-term forecasts
in an autoregressive manner.

\section{Related Work}

Semantic forecasting 
anticipates semantic contents
of an unobserved future image.
Conceptually, this task 
can be factorized into 
RGB forecasting \cite{mathieu16iclr}
and semantic prediction
in the forecasted image.
However, we prefer to carry it out 
as a single processing step 
\cite{luc2017predicting}
due to advantages 
stated in the introduction.
In particular, we consider forecasting 
three dense prediction tasks:
semantic segmentation \cite{zhou2019semantic},
instance segmentation \cite{he2017mask}
and panoptic segmentation \cite{Cheng_2020_CVPR}.
Our method warps features from observed images
into their future positions,
which makes it related
to optical flow \cite{liu19cvpr}.
Our work is most related
to previous 
dense semantic forecasting approaches 
F2F \cite{luc2018predicting} and M2M \cite{terwilliger2019recurrent}.

\subsection{Optical flow}

Optical flow is a dense 2D-motion field
between neighbouring image frames 
$\mathrm{I}_{t}$ and $\mathrm{I}_{t+1}$.
A future image $\mathrm{I}_{t+1}$
can be approximated 
either by forward warping $\mathrm{I}_{t}$
with the forward flow 
$\mathbf{f}_t^\mathrm{t+1}$=
$\textrm{flow}(\mathrm{I}_{t},\mathrm{I}_{t+1})$,
or by backward warping it 
with the backward flow
$\mathbf{f}_{t+1}^{t}$=
$\textrm{flow}(\mathrm{I}_{t+1},\mathrm{I}_{t})$
\cite{szeliski2010computer}:
\begin{align}
  \label{eq:flow}
  \mathrm{I}_{t+1} 
  \approx
    \mathrm{warp\_fw}
      (\mathrm{I_t}, \mathbf{f}_t^{t+1})
  \approx
    \mathrm{warp\_bw}
      (\mathrm{I_t}, \mathbf{f}_{t+1}^{t})
\end{align}
The approximate equality
reminds us that bijective mapping
can not be established 
due to occlusions and disocclusions.

Recent optical flow methods
leverage deep convolutional models \cite{dosovitskiy15iccv,sun18cvpr}
due to capability to guess motion 
where the correspondence 
is absent or ambiguous.
These approaches exploit 
explicit 2D correlation
across the local neighbourhood
\cite{dosovitskiy15iccv},
and local embeddings 
which act as a correspondence metric \cite{zbontar2016stereo}.
Our method is especially related
to self-supervised approaches
\cite{liu19cvpr}
since they learn to reconstruct optical flow
without any groundtruth information.

\subsection{RGB forecasting}

RGB forecasting is also known 
as video prediction
\cite{mathieu16iclr}.
The task is to predict 
one or more future video frames 
given a few recently observed frames
of the same scene
\cite{oprea2020pami}.
This is especially interesting
due to opportunity for self-supervised 
representation learning
on practically unlimited data.

Mathieu et al.\ \cite{mathieu16iclr} express 
RGB forecasting as image generation 
with a multiscale adversarial network.
Vukotic et al.\ \cite{vukotic2017iciap} embed
the desired temporal offset 
in the latent representation
of the observed scenery.
Reda et al.\ \cite{reda2018sdc} warp observed frames 
by applying a regressed kernel 
at the location determined 
by the forecasted flow.

Some works generate 
video from still images 
by warping the image
with forecasted flow
\cite{li2018flow,pan2019video}.
Our approach forecasts multiple flows which warp
multiple previous features towards 
a single future feature tensor.
This allows to resolve some disocclusions
by warping from a suitable past image.

Some works decompose RGB forecasting
into reconstruction from the past
and in-painting the novel scenery
\cite{li2018flow,hao2018controllable,
  gao2019disentangling}.
Their forecast includes
the future warp and the 
respective disocclusion map.
The past frame is first warped
with forecasted flow,
and then the disocclusions 
are filled by in-painting.
This setup is conceptually similar 
to the proposed F2MF approach,
however there are 
two important differences as follows.
First, we exploit spatio-temporal
correlation features
and deformable convolutions.
Second, we perform the forecast 
on heavily subsampled (16$\times$ or 32$\times)$
abstract features instead of pixels.
This allows our method
to operate on megapixel images
and to achieve state-of-the-art
accuracy in semantic forecasting
while incurring only a modest 
computational overhead
with respect to single-frame prediction.

\subsection{Dense semantic prediction}

Several computer vision tasks
address scene understanding 
at the pixel level.
Semantic segmentation \cite{zhou2019semantic}
assigns each pixel 
to a suitable semantic class.
This includes stuff classes 
such as road or vegetation,
as well as object classes such as person or bicycle. 
Instance segmentation \cite{he2017mask}
detects instances of object classes 
and associates them with 
the respective image regions. 
Panoptic segmentation \cite{Cheng_2020_CVPR}
subsumes the previous two tasks
by assigning pixels 
both the semantic class 
and the instance index.
Today, all these tasks are solved by 
fully convolutional models for
dense prediction.

Several succesful architectures 
for dense semantic prediction have 
an asymetric hourglass-shaped structure
consisting of the recognition backbone
and a lean upsampling datapath
\cite{lin2017feature,kreso17cvrsuad,
  Cheng_2020_CVPR}.
The recognition backbone usually corresponds 
to a fully-convolutional portion
of a model designed for image classification.
The role of this component
is to convert the input image 
into a subsampled latent representation.
Most authors exploit 
knowledge transfer from ImageNet 
since ima\-ge-wide supervision
requires much less effort 
than dense semantic supervision.
The upsampling datapath recovers 
dense semantic predictions
by blending semantics of deep features
with location accuracy 
of their shallow counterparts.
The blending is usually implemented 
by means of skip-connecti\-ons 
from the backbone towards the upsampling path.
Empirical studies show advantages
of asymetric designs which complement
deep and thick recognition 
with shallow and thin upsampling
\cite{krevso2020efficient}.
This suggests that recognition requires
much more capacity than guessing the borders
when rough semantics is known.

\hyphenation{bhatta-cha-ryya}

\subsection{Forecasting at the level of semantic predictions (S2S)}
Luc et al.\ \cite{luc2017predicting}
were the first to propose 
direct semantic forecasting.
Their S2S model maps 
past semantic segmentations 
into the future semantic segmentation. 
Bhattacharyya et al.\ \cite{bhattacharyya2018bayesian} 
try to account for multimodal future
with variational inference based on MC dropout,
while conditioning the forecast
on measurements from the vehicle odo\-meter.
Rochan et al.\ \cite{rochan2018future} 
formulate the forecasting in a recurrent fashion 
with shared parameters between each two frames.
Chen et al.\ \cite{chen2019multi} improve their work 
by leveraging deformable convolutions 
and enforcing temporal consistency
between neighbouring feature tensors.
Their atten\-tion-based blending
is related to forward warping
based on pairwise correlation features
\cite{Saric_2020_CVPR}.
However, the forecasting accuracy 
of these approaches is considerably lo\-wer 
than in our ResNet-18 experiments
despite greater forecasting capacity 
and better single-frame performance.
This suggests that ease of correspondence 
and avoiding error pro\-pagation 
may be important for successful forecasting.
Concurrent work \cite{graber2021panoptic} 
forecasts panoptic segmentation
by separately regressing 
things and stuff 
from previous predictions.

\subsection{Flow-based forecasting (M2M)}

Direct semantic forecasting 
requires a lot of training data
due to necessity to learn 
all motion patterns one by one.
This can be improved 
by allowing the forecasting model
to access geometric features 
which reflect 2D motion in the image plane
\cite{jin2017predicting}.
Further development of that idea
brings us to warping the last dense prediction
according to forecasted optical flow. 
A prominent instance of this approach 
can be succinctly described
as motion-to-motion (M2M) forecasting
since it receives optical flows
from three observed frames
and produces the future optical flow.
The corresponding implementation
based on convolutional LSTM
had achi\-eved state-of-the-art 
semantic forecasting accuracy 
prior to our work
\cite{terwilliger2019recurrent}.
This approach is
related to our F2M module 
which also forecasts 
by warping with regressed flow.
However, our F2M module operates 
on abstract convolutional features,
and does not require 
external computational resources and 
additional supervision
for training and evaluating the flow model.
Our approach discourages error propagation
due to end-to-end training
and implies very efficient inference
due to subsampled resolution
and feature sharing between 
motion reconstruction and dense recognition.
Additionally, we take into account 
features from all past four frames
instead of relying only on the last prediction.
This allows our F2M module to detect 
complex disocclusion patterns and 
simply copy from the past where possible.
Further, our module has access 
to raw semantic features 
which are complementary to flow patterns 
\cite{feichtenhofer16cvpr},
and often strongly correlated with future motion
(consider e.g.\ cars vs pedestrians).
Finally, we complement our F2M module 
with pure recognition-based F2F forecasting
which outperforms F2M
on previously unobserved scenery.

\subsection{Feature-level forecasting (F2F)}

Feature-to-feature (F2F) forecasting 
maps past features 
to their future counterparts.
The first F2F approach 
operated on image-wide features from 
a fully connected AlexNet layer
\cite{vondrick2015anticipating}.
Luc et al.\ \cite{luc2018predicting} propose 
dense F2F forecasting by regressing 
all features along the FPN-style 
\cite{lin2017feature} upsampling path.
Further work \cite{sun2019predicting, hu2021apanet} 
improves by using a convolutional 
LSTM module at each level 
of the feature pyramid, and proposing
inter-level connections
for context sharing.
However, forecasting at fine resolution
is computationally expensive
\cite{couprie18eccvw}.
Hence, we propose single-level forecasting 
of the coarsest features 
\cite{vsaric2019single,chiu2020segmenting}.
Such aproach is advantageous
due to small inter-frame displacements,
rich contextual information
and small computational footprint \cite{vsaric2019single}.
It also has an intuitive appeal
of prioritizing the big picture
before moving on to the fine 
details.
Most recent work \cite{lin2021predictive}
follows and extends our idea of 
single-level feature forecasting.
They introduce a variational autoencoder
which provides a compressed representation
of a multi-level feature pyramid.
After the forecasting,
this representation
is decoded into tensors
of the FPN style decoder.

Vora et al.\ \cite{vora2018future} formulate
feature-level forecasting
as reprojection of reconstructed features
to the forecasted future ego-location.
However, such approach 
underperforms in presence of 
(dis-)occlusions and 
lar\-ge cha\-nges of perspective.
Additionally, it makes it difficult to account 
for independent motion of moving objects.
Large empirical advantage of our method
suggests that optimal forecasting performance 
requires a careful balance 
between reconstruction and recognition,
as well as that explicit 3D reasoning 
may bring minor benefits.

\subsection{Semantic forecasting in video}

Dense semantic forecasting has been used
to increase the level of supervision
in partially labeled video \cite{zhu19cvpr}.
This could be especially useful
when only few \cite{robinson20cvpr} 
or weak \cite{zhang20tpami}
labels are available.
Semantic forecasting can
improve inference in video
by alleviating disturbances 
due to noisy input 
\cite{davison07pami}.
Learned object dynamics 
could especially improve 
few-shot approaches
which rely on inference-time optimization
\cite{han18cvpr}.
\section{Dense semantic forecasting with F2MF}

\begin{figure}[h!]
  \centering
  \includegraphics[width=\columnwidth]
      {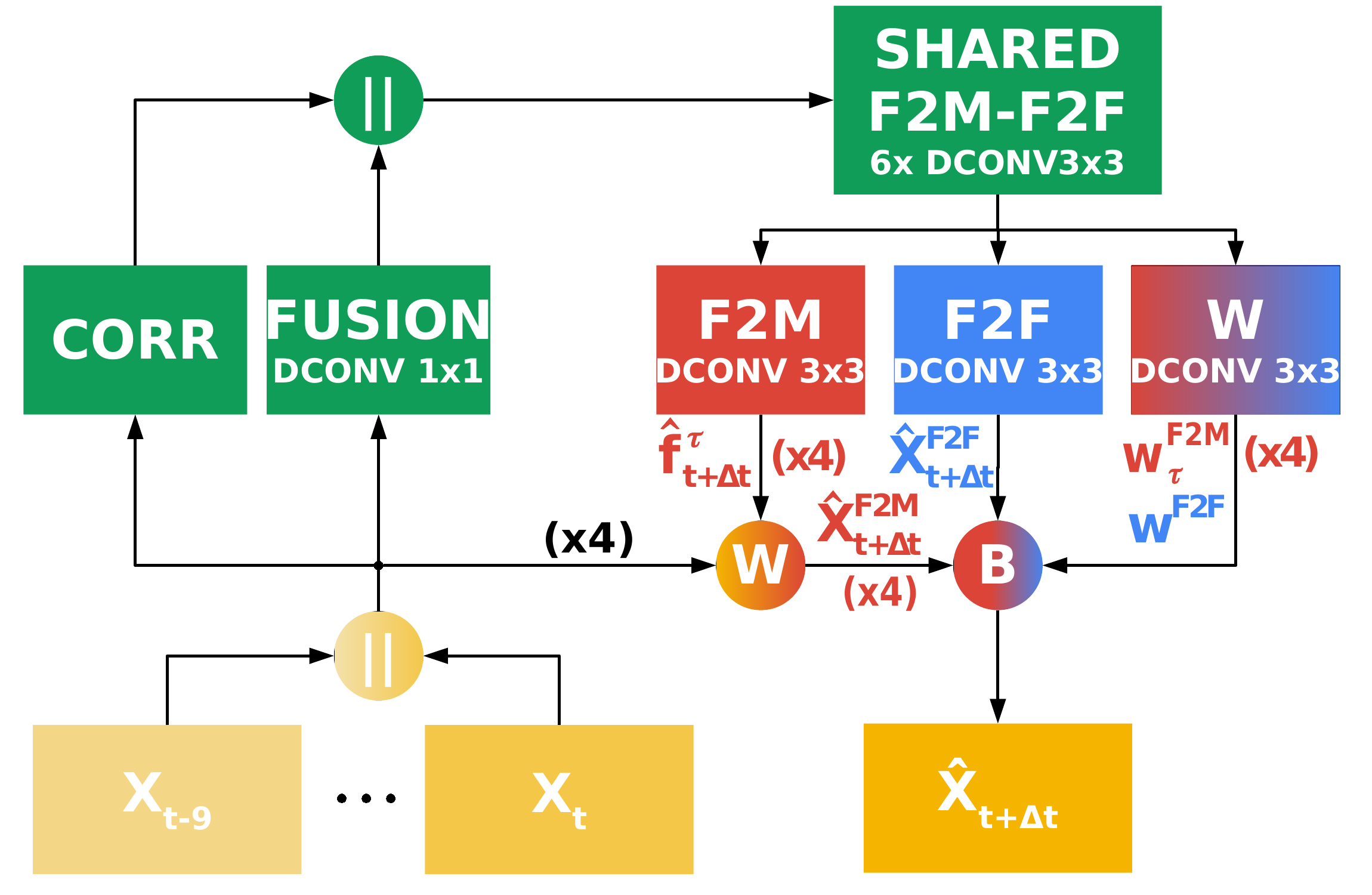}
  \caption{Details of F2MF forecasting.
    F2M and F2F modules receive 
    a processed concatenation ($| |$)
    of features from observed frames
    ($\mathbf{X}_{t-9:t:3}$), 
    and their spatio-temporal 
    correlation coefficients.
    The F2M module regresses 
    future feature flow 
    which warps (W) past features 
    into their future locations. 
    The F2F module forecasts 
    the future features directly.
    The compound forecast 
    $\mathbf{X}_{t+\Delta t}$
    is a weighted blend (B) 
    of F2M and F2F forecasts.
  }
  \label{fig:f2mf_model}
\end{figure}

We present a novel forecasting approach
which combines 
standard feature-level forecasting (F2F)
with its regularized variant (F2M) 
which warps past representations
into the future.
Fig.\ \ref{fig:f2mf_model}
illustrates the structure
of the proposed joint model.
On input, we receive
$T$=$4$ past feature tensors
$\mathbf{X}_{t-9}$,
$\mathbf{X}_{t-6}$,
$\mathbf{X}_{t-3}$, 
$\mathbf{X}_{t}$ 
($\mathbf{X}_\mathtt{t-9:t:3}$ for short)
extracted with the front end
of the desired single-frame model
for dense semantic prediction.
Our F2MF model 
maps the past features
into the future feature tensor
$\mathbf{\hat{X}}_{t+\Delta t}$.
This tensor is transformed 
to semantic predictions $\mathbf{\hat{S}}_{t+\Delta t}$
throughout the back end 
of the chosen single-frame model
as shown in Fig.\ \ref{fig:intro}.
Note that $\Delta t$ denotes 
the temporal offset
between the last observed frame
and our forecast.

F2MF forecasting proceeds as follows. 
Input features are concatenated 
across the semantic dimension
and fused with a single convolutional layer 
in order to reduce the number of channels.
In parallel, the correlation module computes
pairwise correlation coefficients
across a local neighborhood.
The fu\-sed representation
is concatenated with
the correlation features.
The result is further processed
through six convolutional layers
in order to recover 
the shared representation
for F2M and F2F forecasting.
The weight module 
(block W in Fig.~\ref{fig:f2mf_model})
outputs five feature maps: 
$w^\text{F2M}_{t-9}, w^\text{F2M}_{t-6}, w^\text{F2M}_{t-3}, w^\text{F2M}_t$
and $w^\textrm{F2F}$.
The four maps
$[w^\textrm{F2M}_\tau]$,
$\tau\in\{t-9:t:3\}$
represent contributions
of the four past feature tensors
in the F2M forecast.
The fifth map $w^\textrm{F2F}$
represents the contribution 
of the F2F head
in the compound forecast.
All convolutional layers 
are implemented as BN-ReLU-dconv,
where dconv stands 
for deformable convolution
\cite{zhu2018deformable}.

\newcommand{\myX}[1]
  {\mathbf{\hat{X}}_
    {t+\Delta t}^\mathrm{#1}}

\subsection{F2M module}

The F2M module assumes 
that the future 
can be explained 
as a geometrical transformation
of the observed past.
Therefore, it outputs 
a dense field of motion vectors 
$\mathbf{\hat{f}}_{t+\Delta t}^{\tau}$
for each of the T=4 
input feature tensors
$\mathbf{X}_\tau$,
$\tau\in\{t-9,t-6,t-3,t\}$.
The future tensors
${\hat{X}}^{(\tau)}_{t+\Delta t}$
are estimated 
by backward warping
\cite{szeliski2010computer}
the corresponding input features 
$\mathbf{X}_\tau$
with the regressed flow $\mathbf{\hat{f}}_{t+\Delta t}^{\tau}$.
The resulting warped tensors 
are subsequently blended
with regressed weights 
which we activate with per-pixel softmax.
Thus, the F2M forecast is
a weighted sum of warped features 
from the observed images:
\begin{align}
  \quad
  \mathbf{\hat{X}}^{(\tau)}_{t+\Delta t} &= \mathrm{warp\_bw}(
      \mathbf{X}_\tau, 
      \mathbf{\hat{f}}_{t+\Delta t}^{\tau})
  \\
  \myX{F2M} &= 
    \sum_\tau \mathbf{\alpha}_\tau \cdot 
    \mathbf{\hat{X}}^{(\tau)}_{t+\Delta t}
  \label{eq:f2m}
  \\
  \bm{\alpha} &= \textrm{softmax}(
    [w^\textrm{F2M}_\tau]_
      {\tau\in\{t-9:t:3\}})
    \;.
\end{align}
This allows the F2M module to choose
the most suitable previous image
for forecasting a particular region.
Such opportunity is particularly beneficial 
in some occlusion patterns
as we illustrate in Fig.\ \ref{fig:gradients}.

However, the assumption that the future 
can be reconstructed from the past
is only partially true,
since the future often 
brings unpredictable novelty. 
Additionally, sometimes the correspondence 
will be hard to determine
due to large ego-motion 
and changes in perspective.
Consequently, some parts of the scene
are going to be particularly 
hard to forecast 
by simple warping from the past.
Accurate forecasting of such regions 
will require imagination ability 
as we describe next.

\subsection{F2F module}

The F2F module directly regresses 
future features 
$\myX{F2F}$
from the shared representation.
It consists of a single 
BN-ReLU-dconv unit,
where the number of output feature maps
is the same as in 
a single past feature tensor.
As it is not bound to 
reconstruction from the past,
it has a chance to in-paint
the features in novel regions.
This is similar to some previous 
\cite{luc2018predicting}, 
and concurrent work
\cite{sun2019predicting,chiu2020segmenting}, 
however there are three important differences. 
First, we aim at single-level F2F forecasting. 
Second, we use deformable convolutions. 
Third, our F2F module has access 
to spatio-temporal correlation features
which relieve the need 
to learn correspondence from scratch.
Our experiments show clear advantage 
of these novelties.
The contribution of correlation features suggests 
that correspondence is not easily learned
on existing datasets.

\subsection{Compound F2MF forecasting}

We hypothesize that F2M and F2F forecasting 
might be complementary and 
that their combination 
might lead to improved accuracy.
Consequently, our F2MF module 
formulates future features
as a weighted average 
of the F2M and F2F forecasts
(as before, $\tau\in\{t-9:t:3\}$):
\begin{align}
  \myX{F2MF} &= 
    \beta^\textrm{F2F} \cdot \myX{F2F} +
    \sum_\tau \beta^\textrm{F2M}_\tau \cdot \myX{(\tau)}
  \label{eq:f2mf}
  \\
  \bm{\beta} &=
    [\beta^\textrm{F2F}] 
    \mathbin\Vert
    [\beta^\textrm{F2M}_\tau]
    =
   \textrm{softmax}(
    [w^\textrm{F2F}] 
    \mathbin\Vert
    [w^\textrm{F2M}_\tau])
    \;.
\end{align}
This formulation encourages specialization
of the F2F module for the novel parts of the scene.
It also relaxes the penalty
of the F2M module in such regions
and allows it to focus on parts
where correspondence can be established.
Nevertheless, this kind of separation
might weaken the learning signals
within the two individual modules.
Consequently, we propose to train
F2MF forecasting with three loss terms.
The main loss 
$\mathcal{L}_\mathrm{F2MF}$
involves the F2MF forecast
(\ref{eq:f2mf}).
The two auxiliary losses
$\mathcal{L}_\mathrm{F2F}$
and $\mathcal{L}_\mathrm{F2M}$
affect the corresponding 
outputs $\myX{F2F}$
and $\myX{F2M}$.
All three losses are formulated
as mean squared L2 distance 
between the forecast
and actual features computed
with the single-frame model 
in the future frame.
Consequently, the model can be trained
with self-supervision on unlabeled video.

\subsection{Correlation module}

Our correlation module determines
spatio-temporal correspondence
between neighbouring frames.
On input, it receives convolutional features 
$\mathbf{X}_\mathtt{t-9:t:3}$
in the form of a 
T\,$\times$\,C $\times$\,H\,$\times$\,W tensor.
On output, it produces  
spatio-temporal correlation coefficients
across a d$\times$d neighborhood
for each of the T$-$1 pairs 
of neigbouring frames.

The module first embeds 
features from all time instants
into a space with enhanced metric properties
by a shared 3$\times$3$\times$C' convolution
(C'=128).
We hypothesize that this mapping 
might recover distinguishing information
which is not needed for single-frame inference.
Subsequently, we construct our metric embedding 
by normalizing C'-dimensional features to unit norm
so that cosine similarity becomes dot product.
This results in 
a T$\times$C'$\times$H$\times$W 
metric embedding $\mathbf{F}$.
Finally, we produce $d^2$ correspondence maps
between features at time $\tau$
and their counterparts at $\tau-3$
within the local $d \times d$ neighborhood
for each $\tau\in\{t-6,t-3,t\}$.
We usually set $d=9$.
We denote the value of 
the correlation tensor $\mathbf{C}^\tau$
at location $\mathbf{q}$ 
and feature map $ud+v$
as $\mathbf{C}^{\tau}_{ud+v,\mathbf{q}}$.
This value corresponds to a dot product 
between a metric feature at time $\tau$ 
and location 
$\textbf{q}\in\mathcal{D}(\mathbf{F})$,
and its counterpart 
at time $\tau-3$
and location $\textbf{q} + (u,v)$
where $u,v \in 0\,..\,d-1$
\cite{dosovitskiy15iccv}:
\begin{align}
  \label{eq:corr}
  \mathbf{C}^{\tau}_{ud+v,\mathbf{q}} = 
     \mathbf{F}^{\tau\top}_\mathbf{q}
     \mathbf{F}^{\tau-3}_{\mathbf{q}+[u - \frac{d}{2}, v - \frac{d}{2}]}
    ,
    \mathrm{where}\  
    u\mathrm{,}v \in 
    [0\,..\,d).
\end{align}
Each of the $d^2$ feature maps of
$\mathbf{C}^{\tau}$
can be computed
by elementwise multiplication
of $\mathbf{F}^{\tau}$ and
shifted $\mathbf{F}^{\tau-3}$
followed by channel reduction.
Please note that 
$\mathcal{D}(\mathbf{F})$
denotes the set 
of all spatial locations
in $\mathbf{F}$: 
$\mathcal{D}(\mathbf{F})$=
 \{1..H-1\}$\times$\{1..W-1\}.

\subsection{Application to 
  popular single-frame models }

Figure \ref{fig:intro} suggests that
our F2MF approach can be applied
to a variety of single-frame architectures
for dense semantic forecasting.
The pipeline consists of three processing steps.
First, the front end of the single-frame model 
extracts features from multiple past frames.
Second, our F2MF module 
processes the present features
and recovers the forecast
of the future features.
Finally, the back end 
of the single-frame model
processes the forecasted features 
and forms the 
future dense semantic prediction.

Different than \cite{luc2018predicting},
the presented forecasting approach
addresses features at only one level of abstraction
\cite{vsaric2019single}.
We propose to forecast succinct abstract features
in order to encourage the forecasting module
to focus on the big picture.
This approach is also advantageous
since low spatial resolution 
makes it easier to establish 
spatio-temporal correspondence.

However, ladder-style architectures 
\cite{kreso17cvrsuad, lin2017feature}
present challenges towards 
implementing single-level forecasting.
Their skip-connections require 
either multi-level forecasting
\cite{luc2018predicting}
or single-level forecasting 
of high-resolution features
from the last upsampling module.
Unfortunately, both of these options
have to confront high computational complexity
and large displacements.
None of our experiments along these directions
outperformed our single-level designs.
We therefore base our experiments
on single-frame architectures 
with a reduced number of skip connections, 
which leads to faster training and inference.
This reduces
single-frame performance on small objects,
however the forecasting penalty 
will be contained since small objects
are the hardest to forecast.

\subsection{Forecasting semantic segmentation}

\begin{figure}[htb]
    \centering
    \includegraphics[width=\columnwidth]{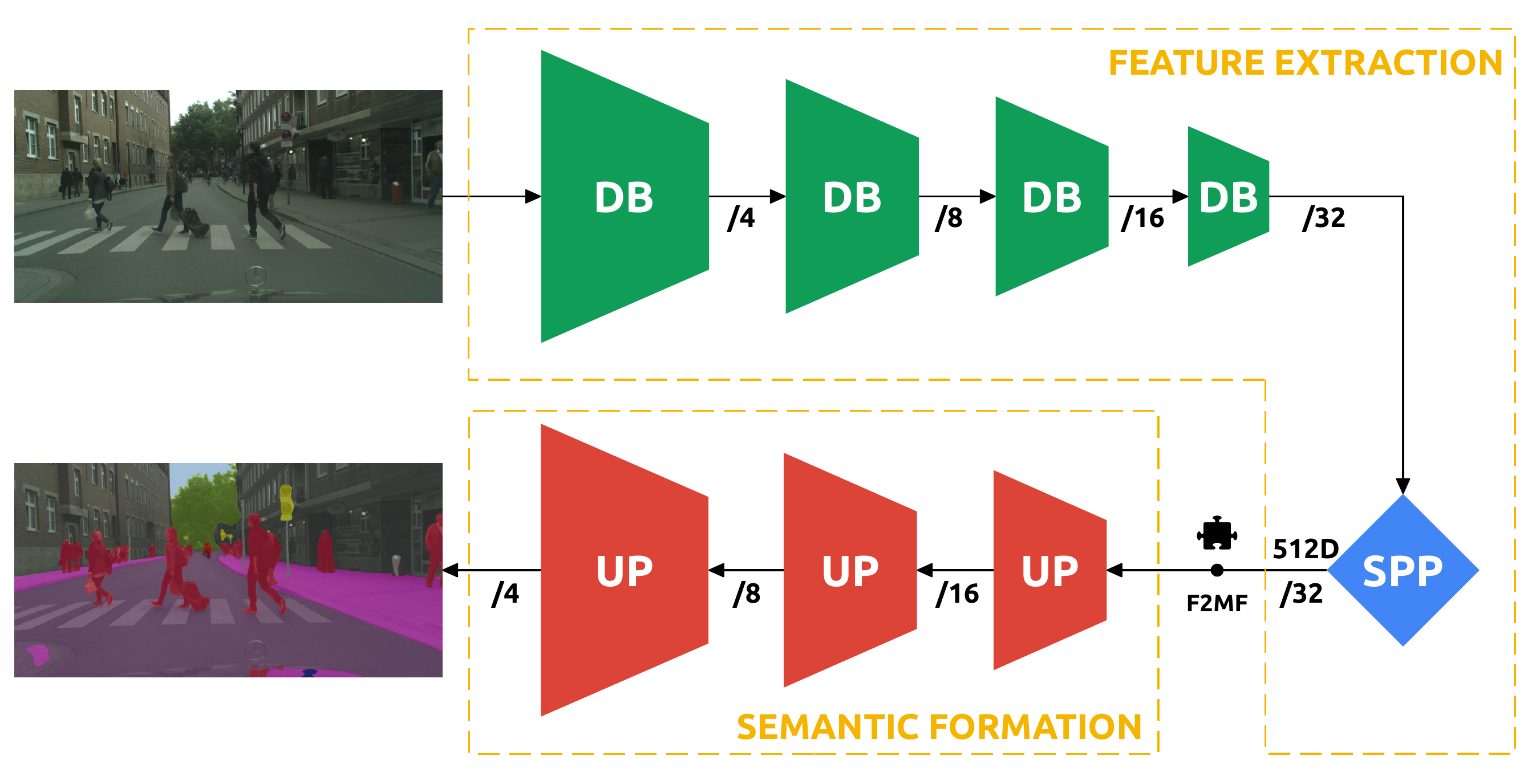}
    \caption{
        Details of our single-frame 
        model for semantic segmentation.
        Notice the absence of skip connections
        from the 
        backbone (green)
        towards the upsampling path (red).
        The black jigsaw 
        symbol
        indicates the feature tensor
        which we use for F2MF forecasting. 
        Feature extraction and semantic formation 
        modules are in correspondence
        with Fig.\ \ref{fig:intro}.
    }
    \label{fig:semseg}
\end{figure}

In the case of semantic segmentation
we achieve our best results 
with custom single-frame models 
without any skip connections.
The single-frame penalty 
of this design decision is 
around 3 percentage points (pp) mIoU
on Cityscapes val.
Despite this handicap, 
our best mid-term forecasting accuracy
outperforms the state-of-the-art
for more than 6\,pp mIoU.
Figure \ref{fig:semseg} shows 
that our F2MF forecasting
addresses 512-dimensional features 
produced by the SPP module,
as well as that the semantic forecast
is formed by the upsampling path
of the single-frame model.

\subsection{Forecasting instance segmentation}

\begin{figure}[h]
  \centering
  \includegraphics[width=\columnwidth]{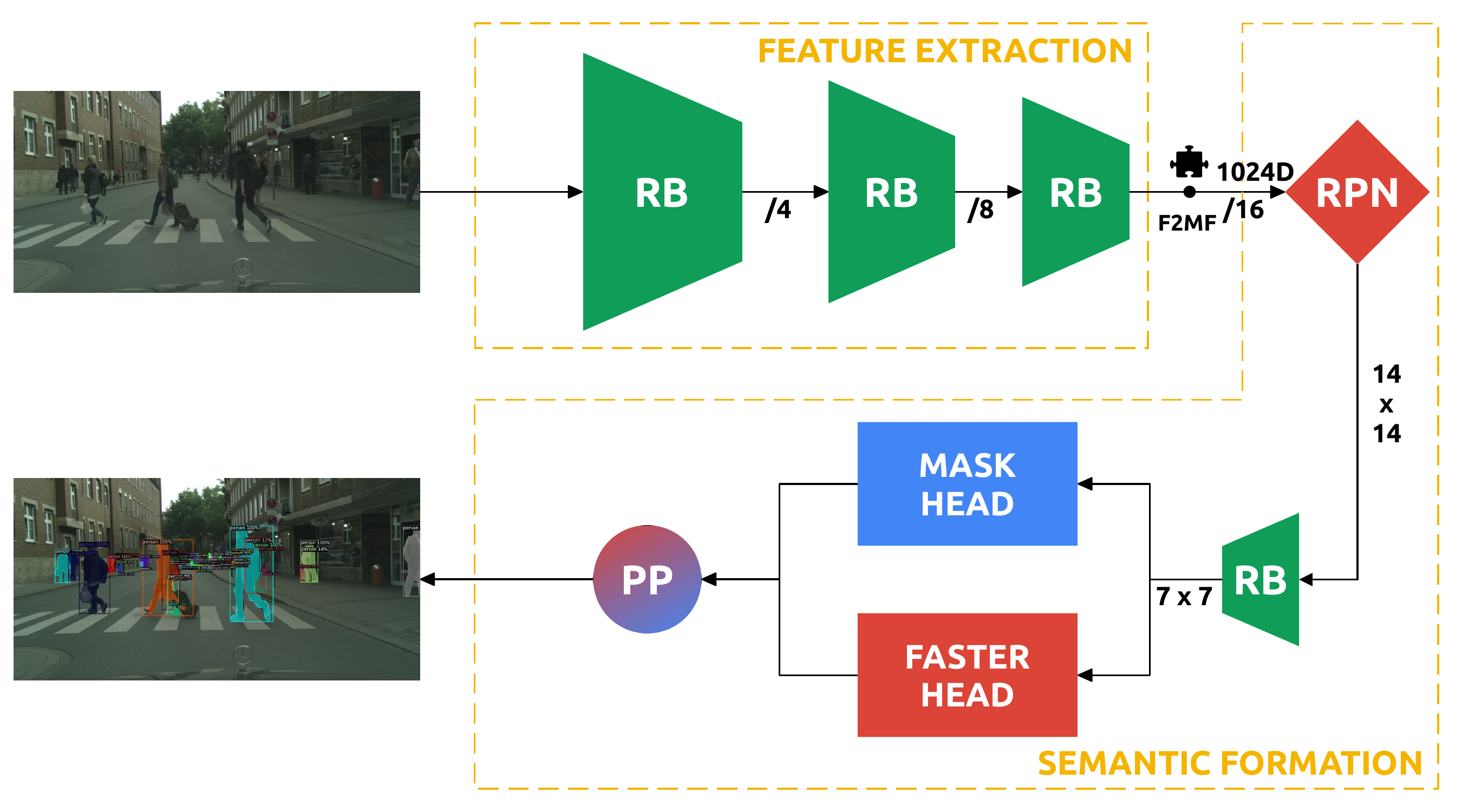}
  \caption{Details of a single-frame model
    which we use in instance segmentation experiments.
    This is a Mask R-CNN model \cite{he2017mask}
    which generates proposals and detections 
    at 16$\times$ subsampled resolution.
    The black jigsaw 
    indicates the feature tensor
    which we use for F2MF forecasting. 
    Feature extraction and semantic formation 
    modules are in correspondence
    with Fig.\ \ref{fig:intro}.
}
    \label{fig:mask}
\end{figure}

In the case of instance segmentation,
we choose a third-party implementation
of a Mask R-CNN
without skip connections
\cite{he2017mask}.
The single-frame penalty 
of this design decision is 
around 1\,pp AP COCO
and 2.7\,pp AP50
on Cityscapes val.
Despite this handicap, 
our model performs favourably
with respect to the state-of-the-art.
Figure \ref{fig:mask} shows 
that our F2MF forecasting
addresses 1024-dimen\-sional features 
produced by the second-to-last
residual block of the ResNet-50 backbone.
Subsequently, the RPN head 
extracts object candidates 
from the forecasted features.
Instance segmentations are obtained
by processing each object candidate
with the last residual block
and the two Mask R-CNN heads.
Future semantics is formed
by resizing the inferred instance segmentations 
to input resolution.

\subsection{Forecasting panoptic segmentation}

\begin{figure}[h]
  \centering
  \includegraphics[width=\columnwidth]{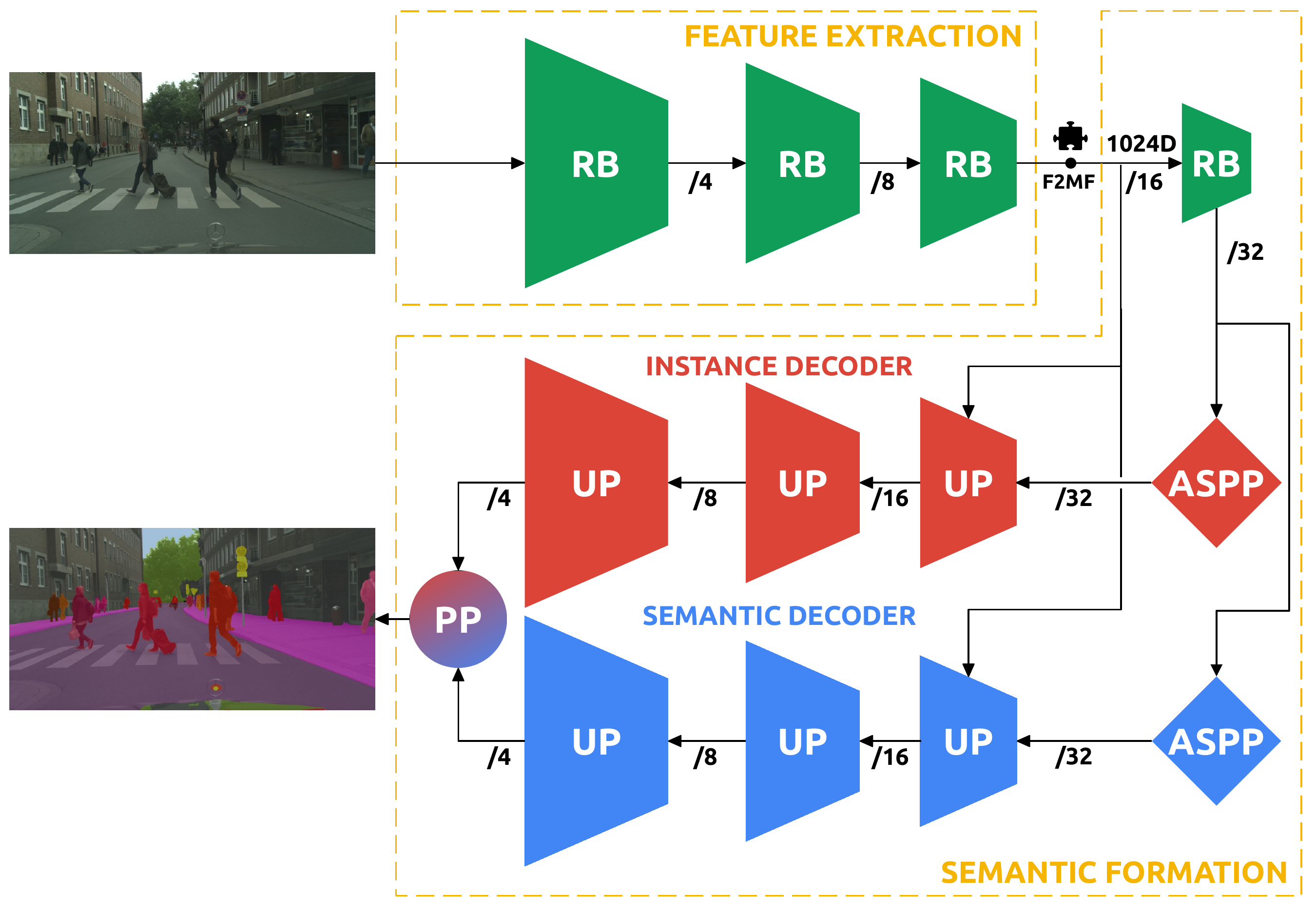}
  \caption{The adopted single-frame panoptic model
    is custom Panoptic DeepLab
    \cite{Cheng_2020_CVPR}
    with only one skip connection.
    This allows us to apply single-level forecasting
    to the feature tensor
    produced by the penultimate residual block,
    as indicated by the 
    jigsaw symbol. 
    Feature extraction and semantic formation 
    modules are in correspondence
    with Fig.\ \ref{fig:intro}.
    }
    \label{fig:pdl}
\end{figure}

In the case of panoptic segmentation,
we train a custom Panoptic DeepLab 
model \cite{Cheng_2020_CVPR}
with only one skip connection.
The single-frame penalty 
of this design decision is 
around 2.3\,pp PQ
on Cityscapes val.
Figure \ref{fig:pdl} shows 
that our F2MF forecasting
addresses 1024-dimensional features 
produced by the 
second-to-last
residual block of the ResNet-50 backbone.
Subsequently, the forecasted features
are processed by the last residual block
and by the two upsampling paths.
Future semantics is finally formed
by postprocessing object centers
and per-pixel semantic 
information.

\section{Experiments}
We train our method on 
Cityscapes train \cite{cordts2016cityscapes},
and evaluate on Cityscapes val and test 
and CamVid test.
We consider short-term and 
mid-term forecasting experiments 
which target 
$\Delta t$=3 and $\Delta t$=9 
time-steps into the future 
(180 and 540 ms respectively)
\cite{luc2017predicting}.
We estimate the forecasting performance
by computing the usual dense prediction 
metrics in the future frame.
We use 
mean intersection over union (mIoU)
for semantic segmentation,
COCO average precision (AP) 
for instance segmentation,
as well as panoptic, segmentation 
and recognition quality 
(PQ, SQ, RQ)
for panoptic segmentation.

Our training procedure involves
several separate steps.
First, the backbone of 
a single-frame model
is pretrained on ImageNet.
Second, we train the 
single-frame model 
on labeled images
in the task-specific setup.
Finally, we train the forecasting module
on unlabeled images 
to map the past features
to their future counterparts
with MSE loss.
We extract all training features 
(both the past and the future) 
by applying the single-frame model 
to the corresponding video frames.
Note that the forecasting module 
does not require any annotations 
in spite of being trained 
in a supervised fashion.

We train our model for 160 epochs
with  ADAM \cite{kingma2014adam} optimizer
and early stopping.
We set the initial 
learning rate to $4\cdot 10^{-4}$ 
and reduce it to $10^{-7}$
through cosine annealing. 
We implement our model in Pytorch.
Training with cached features
takes 12 hours on a single GTX1080Ti.
Training with online data augmentation
requires 2 days and three GPUs.
During training, we normalize inputs 
and outputs of our F2MF model 
to dataset-wide 
zero mean and unit variance. 
During inference, we normalize the input 
and denormalize the output.
This facilitates the training process
and improves generalization.
We set the weights of 
all three components 
of the F2MF loss to 1.
Our data augmentation policy includes
sliding the input tuple across the video clip
and horizontal flipping. 
We refrain from using any auxiliary information 
such as vehicle odometry 
or depth maps.

\subsection{Semantic segmentation forecasting on Cityscapes}
We experiment with 
two single-frame models 
of different capacity.
Both of them have 
an encoder-decoder architecture \cite{orsic2019defense}
without skip connections.
The encoder corresponds to standard
ImageNet classification models.
The decoder consists of 
a spatial pyramid pooling 
and three upsampling modules
as shown in Figure \ref{fig:semseg}.
The two models differ 
in backbone architectures 
and in the width of the decoder.
The weaker model is based on a ResNet-18
and uses 128 feature maps along the upsampling path.
It achieves 72.5\% mIoU on Cityscapes val.
The stronger model uses a DenseNet-121 backbone, 
512 feature maps in the SPP, 
256 maps in the first upsampling module, 
and 128 maps in the last two upsampling modules. 
It achieves 75.8\% mIoU on Cityscapes val.

Table \ref{tab:semseg} shows the forecasting accuracy
for semantic segmentation on Cityscapes val.
The top section shows the performance
of our single-frame model 
which we denote as \emph{oracle} 
since our forecasting models 
generate predictions
for \emph{unobserved} frames.
We also show the performance of 
a simple baseline approach
which copies the segmentation 
from the last observed input frame.
The middle section shows 
experiments from the literature
\cite{luc2017predicting,luc2018predicting,bhattacharyya2018bayesian,rochan2018future,chen2019multi,terwilliger2019recurrent,chiu2020segmenting,vora2018future}.
The bottom section presents performance 
of our F2MF models 
with different single-frame models
and data augmentation policies.
Our best model achieves
state-of-the-art forecasting performance 
with 69.6\% mIoU at short-term 
and 57.9\% mIoU  
at mid-term.
\begin{table}[h]
\caption{
  Evaluation of our F2MF model
  for semantic segmentation forecasting 
  on Cityscapes val.
  \emph{All} denotes all classes,
  \emph{MO} --- moving objects,
  \emph{d.a.}\ --- data augmentation,
  and \dag --- test set.
}
\begin{center}
{
\begin{tabular}{lcccccc}
\toprule
  & \multicolumn{3}{c@{\;}}{Short term:s$\Delta t$=3} & 
    \multicolumn{3}{c}{Mid term: $\Delta t$=9} 
  \\
  Accuracy (mIoU)  &&
  All & MO && All & MO \\
  \midrule
  Oracle-DN121 &&
  75.8 & 75.2 && 75.8 & 75.2 \\
  Oracle-RN18 &&
  72.5 & 71.5 && 72.5 & 71.5 \\
  Copy last (DN121) &&
  53.3 & 48.7 && 39.1 & 29.7 \\
  \midrule
  3Dconv-F2F 
  \cite{chiu2020segmenting}
  &&
    57.0& / && 40.8 & / 
  \\
  Dil10-S2S 
  \cite{luc2017predicting} 
  &&
    59.4 & 55.3 && 47.8 & 40.8 
  \\
  LSTM S2S 
  \cite{rochan2018future}
  &&
    60.1 & / && / & / 
  \\
  Mask-F2F 
  \cite{luc2018predicting} 
  &&
    / & 61.2 && / & 41.2 
  \\
  FeatReproj3D 
  \cite{vora2018future} 
  &&
    61.5 & / && 45.4 & / 
  \\
  Bayesian S2S 
  \cite{bhattacharyya2018bayesian} 
  &&
    65.1 & / && 51.2 & / 
  \\
  LSTM AM S2S 
  \cite{chen2019multi}
  &&
    65.8 & / && 51.3 & / 
  \\
  LSTM M2M 
  \cite{terwilliger2019recurrent} 
  &&
    67.1 & 65.1 && 51.5 & 46.3
  \\ 
  \midrule
  F2MF-RN18 w/o d.a. && 
    66.9 & 65.6 && 55.9 & 52.4 \\
  F2MF-DN121 w/o d.a. &&
    68.7 & 66.8 && 56.8 & 53.1 \\
  F2MF-DN121 w/ d.a. &&
    \textbf{69.6} & \textbf{67.7} &&
    \textbf{57.9} & \textbf{54.6} \\
  F2MF-DN121 w/ d.a. \dag &&
    \textbf{70.2} & \textbf{68.7} &&
    \textbf{59.1} & \textbf{56.3} \\
\bottomrule
\end{tabular}
}
\end{center}
\label{tab:semseg}
\end{table}
The table suggests that
better single-frame model leads 
to better forecasting,
however the single-frame advantage 
decreases in the forecasting setup.
The difference in the oracle performance 
is 3.3 mIoU points, 
but it drops to 1.8 mIoU points at short-term, 
and 0.9 mIoU points in mid-term forecasting.
We retrain our best model 
on trainval and 
submit the test set predictions 
to the online benchmark. 
This resulted in 70.2\% mIoU at short-term 
and 59.1\% mIoU at mid-term forecasting.
This suggests that our performance 
in Table~\ref{tab:semseg} has not been artificially 
improved through validation experiments 
on Cityscapes val.

\subsection{Instance segmentation forecasting on Cityscapes}
Table~\ref{tab:instseg} presents the accuracy 
of our forecasting approach 
on the instance segmentation task  
and compares it to the related work 
on Cityscapes val. 
As in the case of semantic segmentation, 
we use single-level forecasting 
in order to reduce the memory footprint 
and to improve the speed of training. 
We demonstrate that our approach 
is a plug-in solution for any 
dense-prediction model
without skip connections,
by experimenting on a 
public Mask R-CNN model \cite{wu2019detectron2}
which we do not customize (cf.\ Fig.\ 4).
The chosen model (R50-C4) attaches 
the RPN prediction head
to $16\times$ subsampled features 
produced by the penultimate 
residual block of ResNet-50.
We compare our approach with 
previous forecasting approaches \cite{luc2018predicting,sun2019predicting,hu2021apanet}
which use Mask R-CNN 
with FPN upsampling \cite{lin2017feature}.
The two previous approaches 
require forecasting all four levels
of the feature pyramid.
The table shows that our approach
achieves the best short-term accuracy
in spite of considerably weaker oracle
and significantly lower
computational complexity. 
The F2F-Corr baseline
ablates the F2M module
from our design.
The resulting performance
reveals a significant contribution
of F2M forecasting in the case of
instance segmentation.
\begin{table}[h]
\caption{
    Instance segmentation forecasting 
    on Cityscapes val.
}
\begin{center}
{
\begin{tabular}{l@{}cccccc}
\toprule
  & \multicolumn{3}{c@{\;}}{Short term: $\Delta t$=3} & 
    \multicolumn{3}{c}{Mid term: $\Delta t$=9} 
  \\
  &&
  AP & AP50 && AP & AP50 \\
  \midrule
  Oracle (ours) && 
    36.3 & 63.1 && 36.3 & 63.1 \\
  Oracle FPN \cite{luc2018predicting, sun2019predicting, hu2021apanet}&&
    37.3 & 65.8 && 37.3 & 65.8 \\
  Copy last && 
    9.6 & 22.9 && 2.2 & 8.1 \\
  \midrule
  F2F 4$\times$ \cite{luc2018predicting} && 
    19.4 & 39.9 && 7.7 & 19.4  \\
  ConvLSTM F2F 4$\times$ \cite{sun2019predicting} && 
    22.1 & 44.3 && 11.2 & 25.6 \\
  ApaNet \cite{hu2021apanet} &&  23.2 & 46.1 && \textbf{12.9} & \textbf{29.2} \\
  \midrule
  F2F-Corr (ours) && 
    21.2 & 43.3 && 9.4 & 19.2 \\
  F2MF (ours)&& 
    \textbf{23.6} & \textbf{47.2} && 11.5 & 24.2 \\
\bottomrule
\end{tabular}
}
\end{center}
\label{tab:instseg}
\end{table}

\subsection{Panoptic segmentation forecasting on Cityscapes}

Table~\ref{tab:panseg} presents our performance 
on the panoptic segmentation task.
Our single-frame model is 
a custom Panoptic Deeplab \cite{Cheng_2020_CVPR}
with ResNet-50 backbone and 
a single skip connection from the backbone
to the decoder/upsampling path (cf.\ Fig.\ 5).
As in the case of instance segmentation,
we target the features 
at $16\times$ subsampled resolution
from the penultimate residual block.
To the best of our knowledge,
this is the first attempt in forecasting
panoptic segmentation.
The table clearly shows
that our F2MF model
outperforms the copy-last baseline 
by a large margin.
Performance of the F2F-Corr baseline
reveals a significant contribution
of F2M forecasting in the case of
panoptic segmentation.
\begin{table}[h]
\caption{
    Panoptic segmentation forecasting on Cityscapes val.
}
\begin{center}
{
\begin{tabular}{l@{}cccccccc}
\toprule
  & \multicolumn{4}{c@{\;}}{Short term: $\Delta t$=3} & 
    \multicolumn{4}{c@{\;}}{Mid term: $\Delta t$=9} 
  \\
  &&
  PQ & SQ & RQ && PQ & SQ & RQ \\
  \midrule
  Oracle (ours) &&
  57.5 & 79.7 & 70.8 && 57.5 & 79.7 & 70.8 \\
  Oracle (PDL) \cite{Cheng_2020_CVPR} &&
  59.8 & 80.0 & 73.5 && 59.8 & 80.0 & 73.5 \\
  Copy-last && 
  32.3 & 70.9 & 42.4 && 22.3 & 68.1 & 30.2 \\
  \midrule
  F2F-Corr (ours) && 
  43.3 & 74.4 & 55.5 && 27.5 & 69.7 & 36.0 \\ 
  F2MF (ours) && 
  \textbf{47.3} & \textbf{75.1} & \textbf{60.6} && \textbf{33.1} & \textbf{71.3} & \textbf{43.3} \\ 
\bottomrule
\end{tabular}
}
\end{center}
\label{tab:panseg}
\end{table}

Tables I-III show that F2MF forecasting
can be applied to three different
dense semantic prediction tasks
and therefore confirm that 
our method indeed is task-agnostic.

\subsection{Qualitative results}
\newcommand{\myqwidth}{0.25\textwidth}
\begin{figure*}[h]
    \centering
    \begin{tabular}{@{}c@{\,}c@{\,}c@{\,}c@{}}
         \includegraphics[width=\myqwidth]{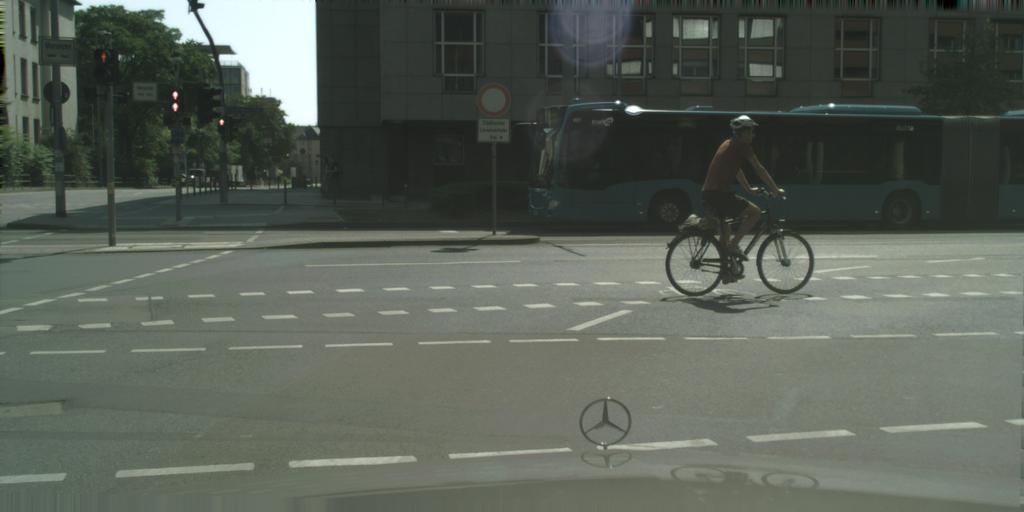} &
         \includegraphics[width=\myqwidth]{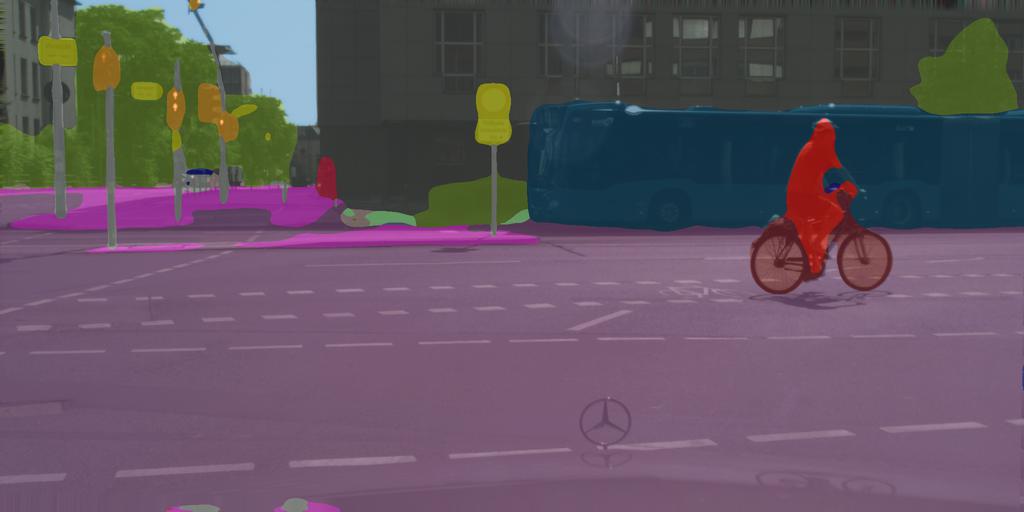} &
         \includegraphics[width=\myqwidth]{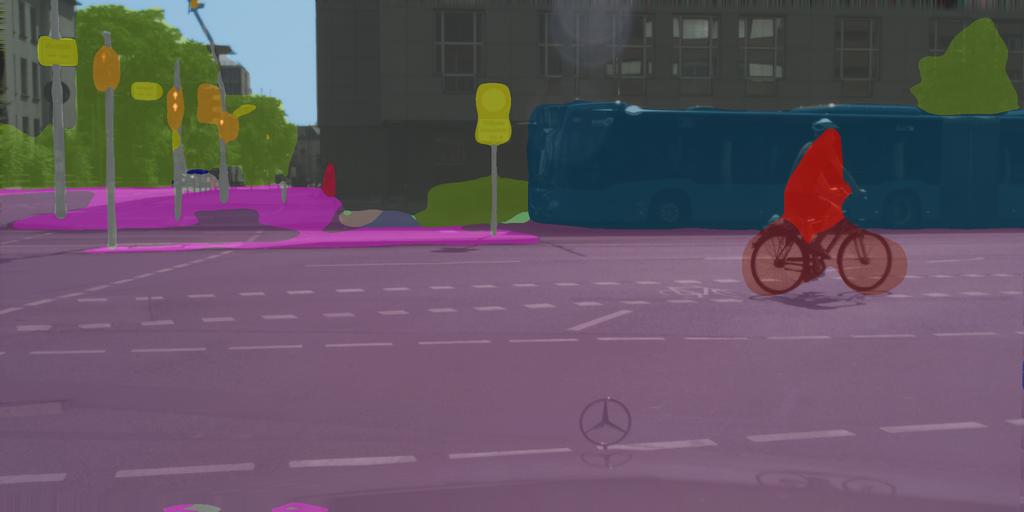} &
         \includegraphics[width=\myqwidth]{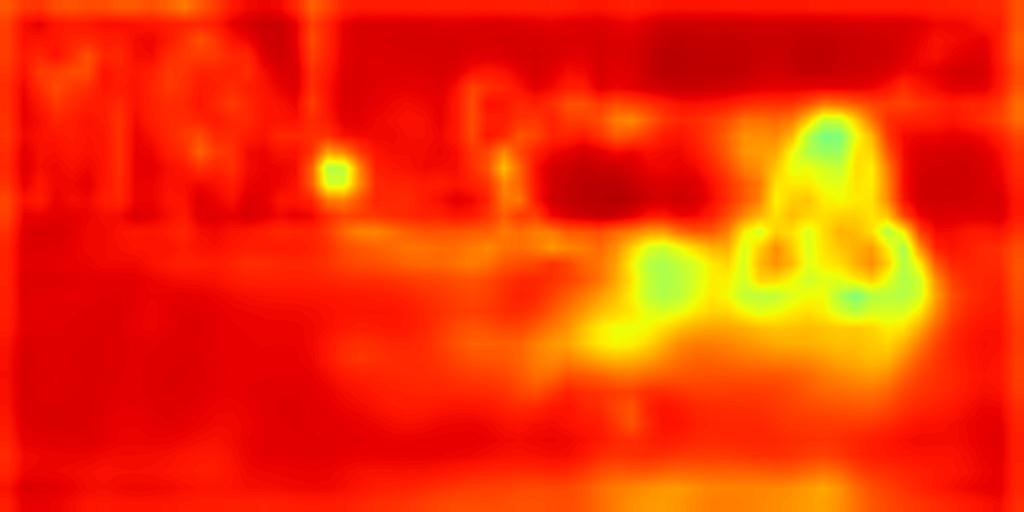} \\ [-0.1em]
        \includegraphics[width=\myqwidth]{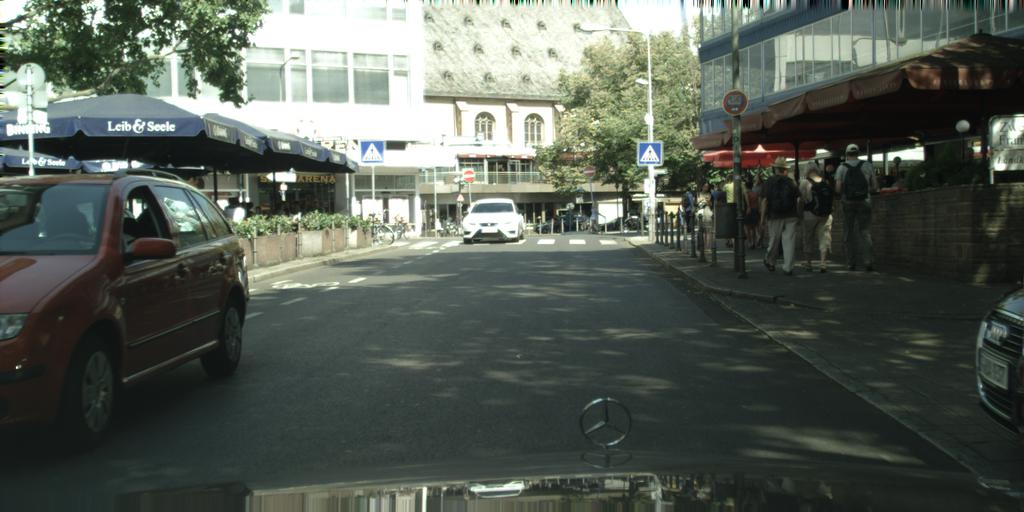} &
        \includegraphics[width=\myqwidth]{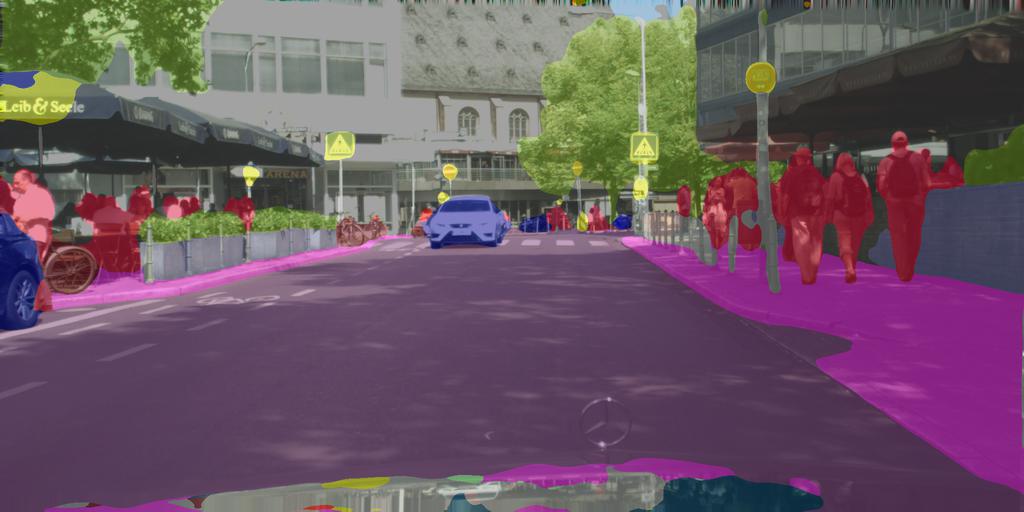} &
        \includegraphics[width=\myqwidth]{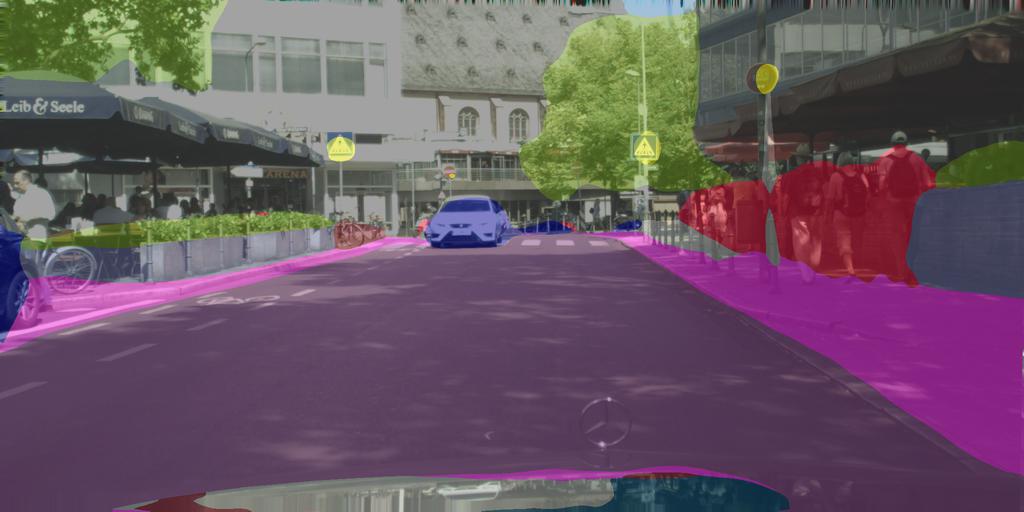} &
        \includegraphics[width=\myqwidth]{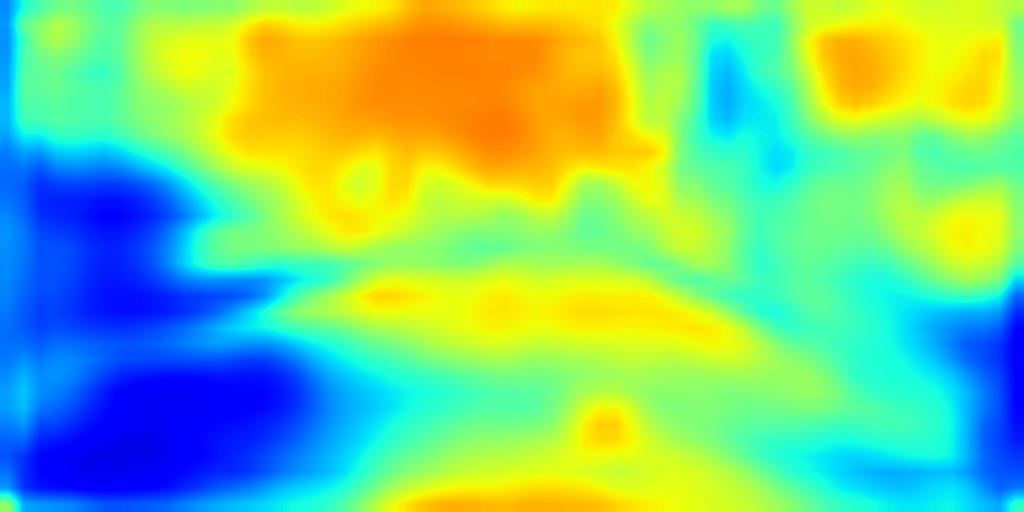} \\ [0.2em]
        
        \includegraphics[width=\myqwidth]{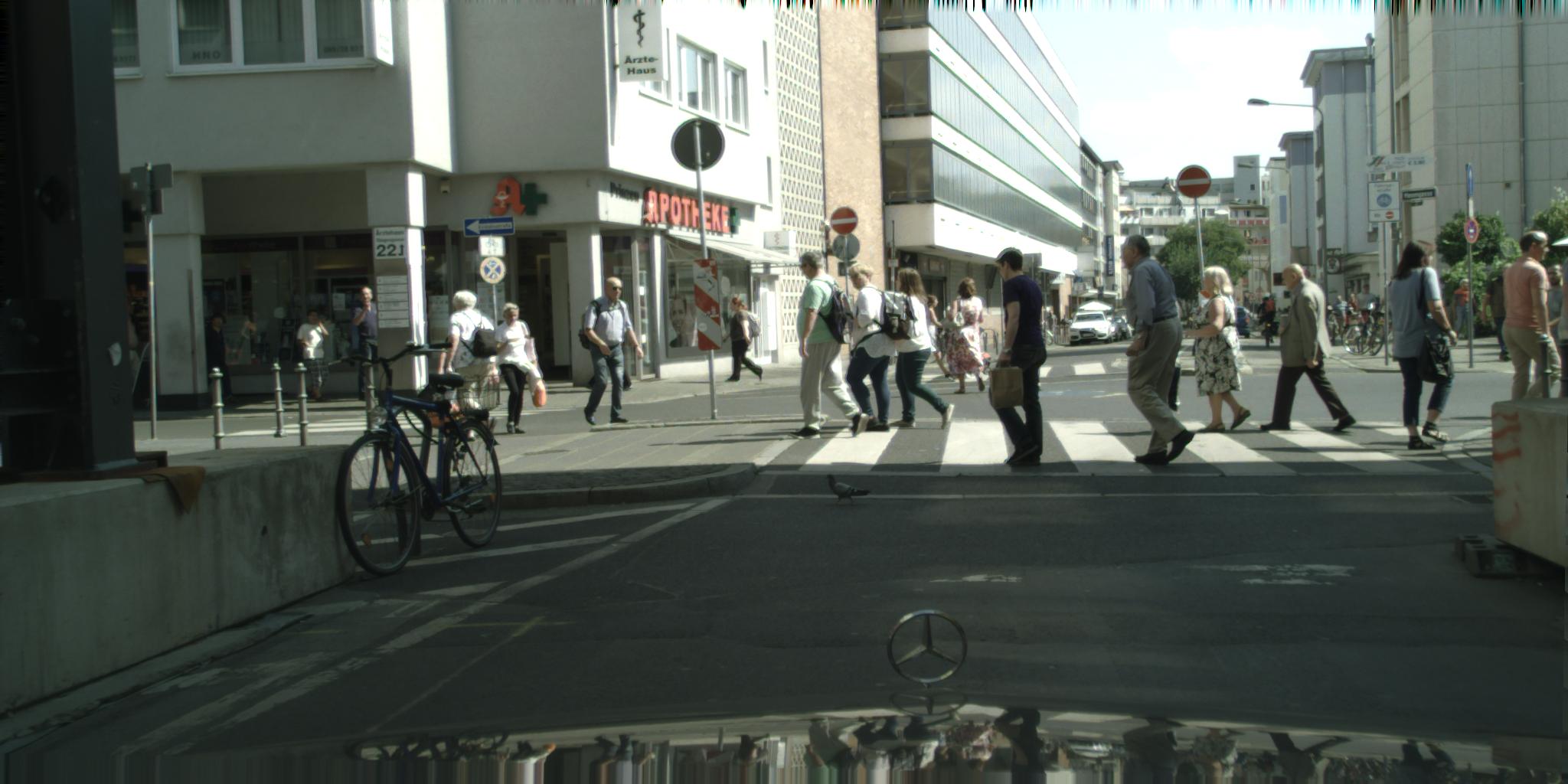} &
         \includegraphics[width=\myqwidth]{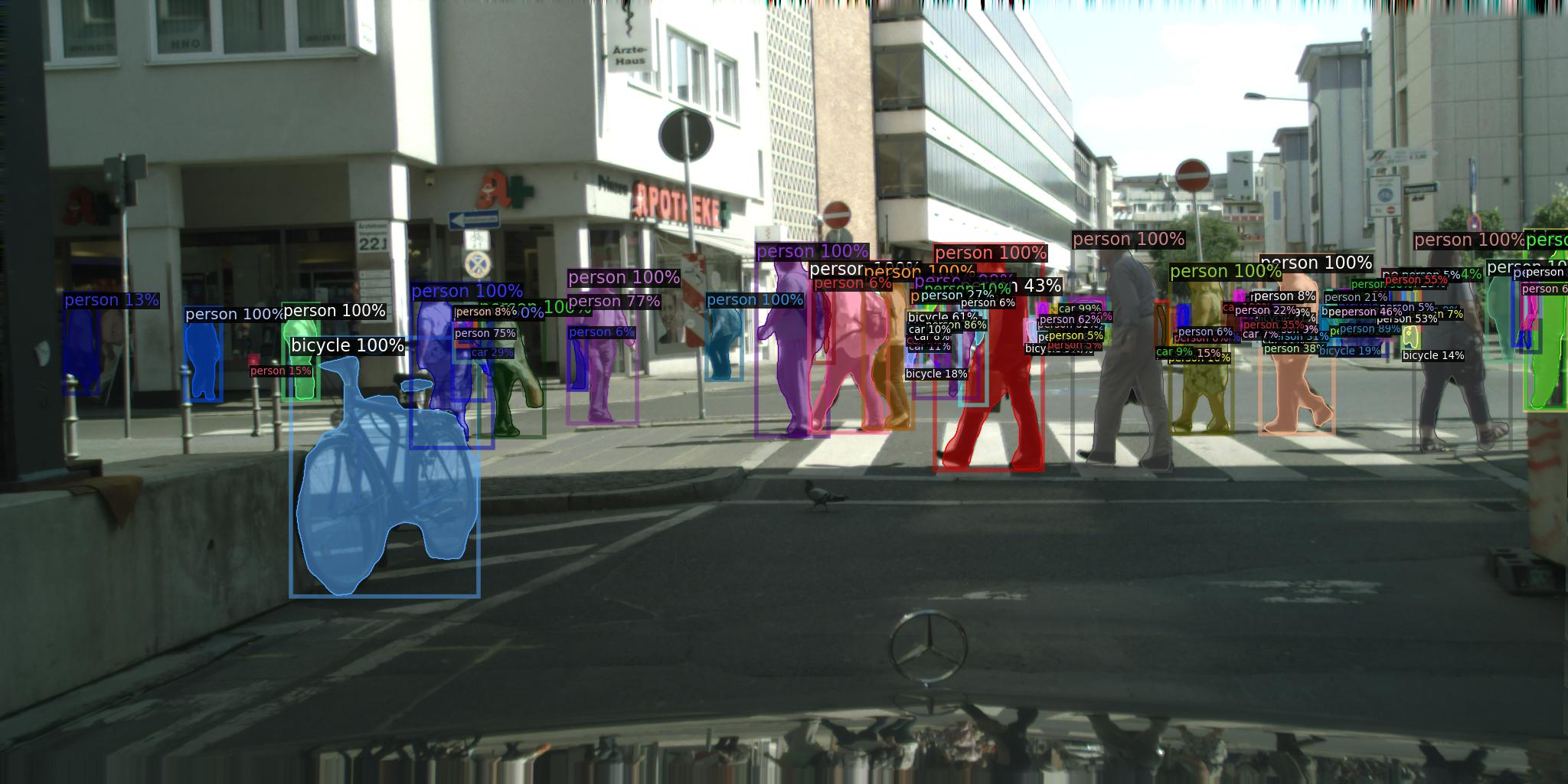} &
         \includegraphics[width=\myqwidth]{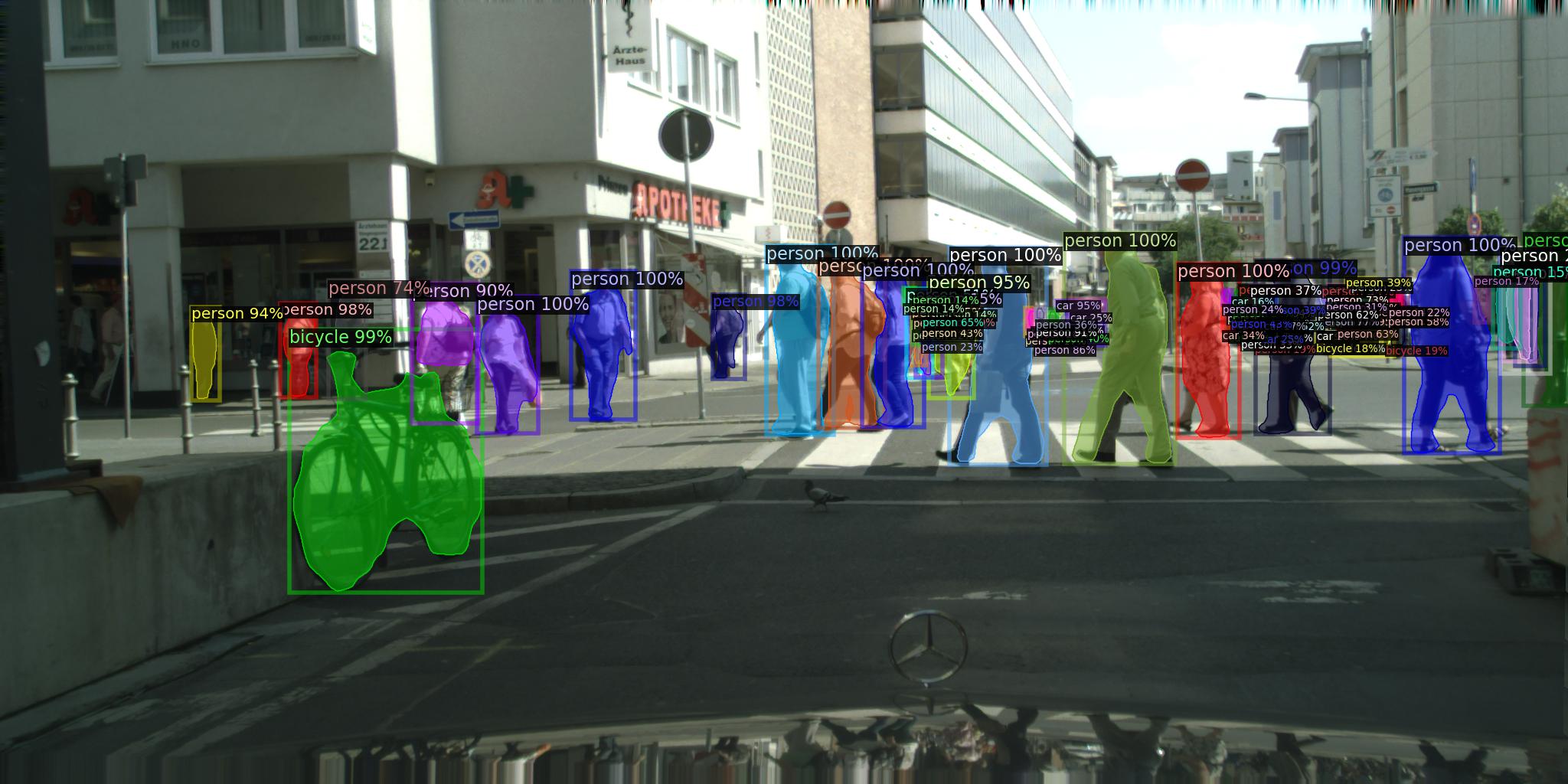} &
         \includegraphics[width=\myqwidth]{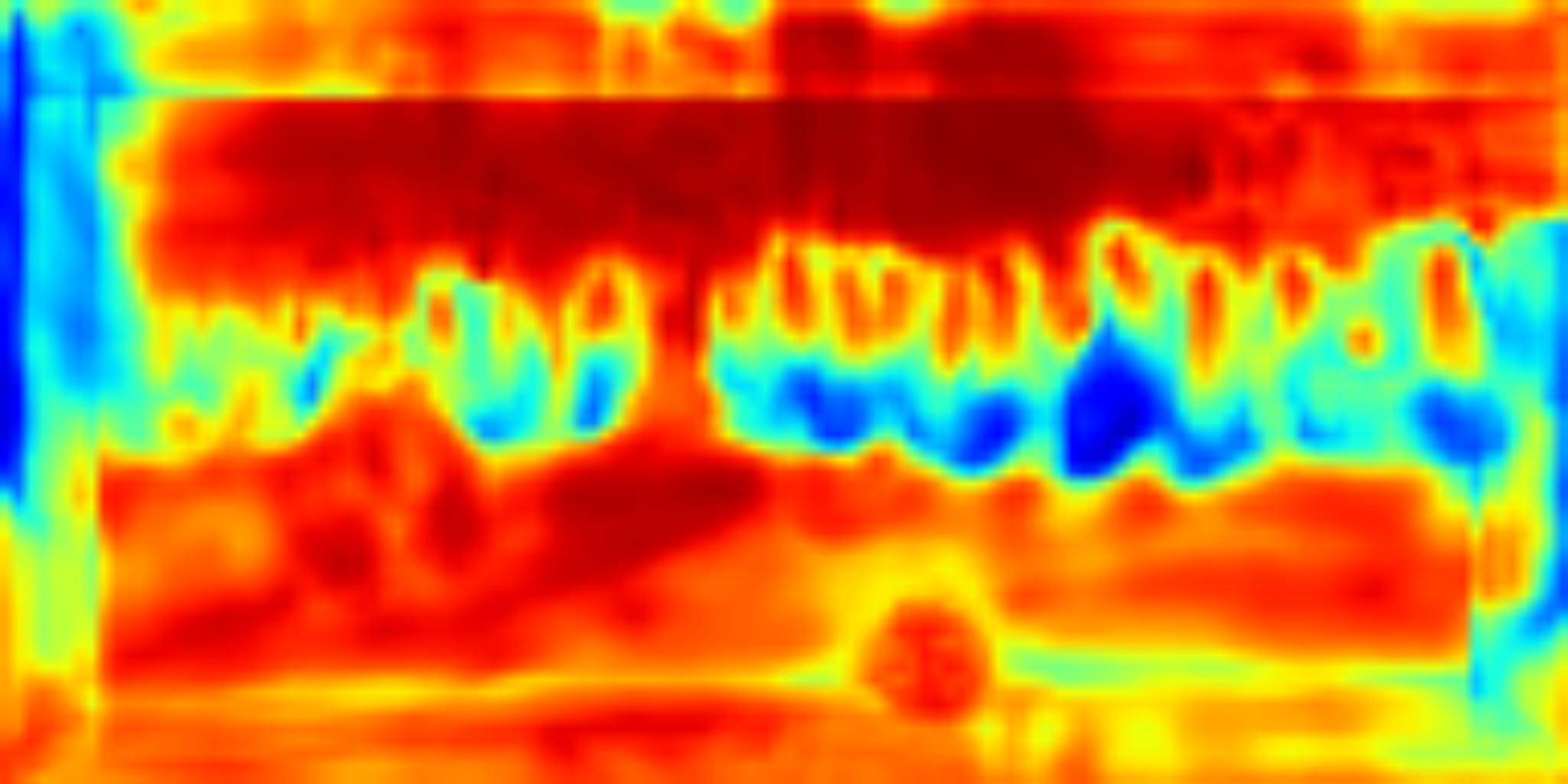} \\[-0.1em]
        \includegraphics[width=\myqwidth]{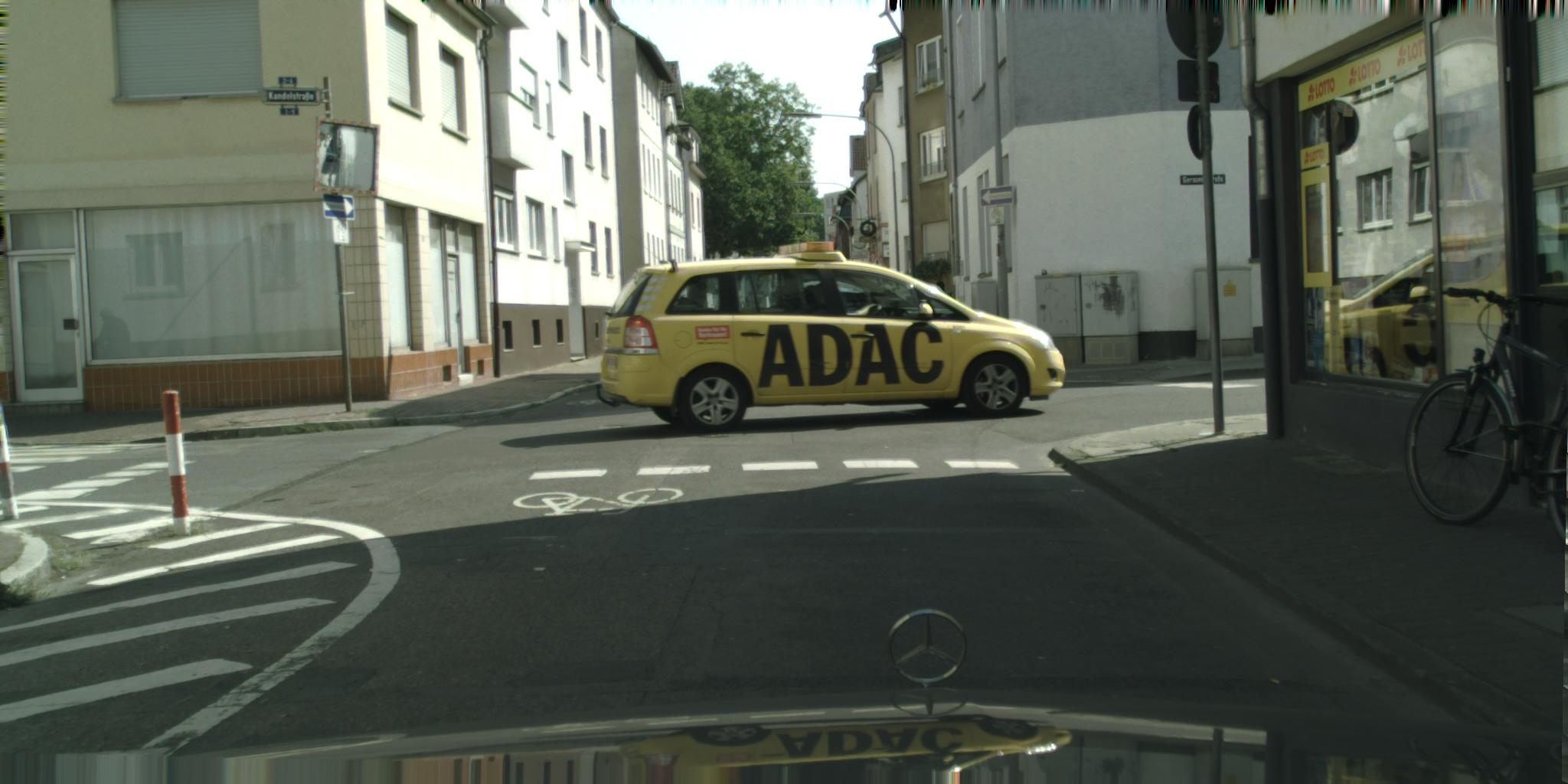} &
        \includegraphics[width=\myqwidth]{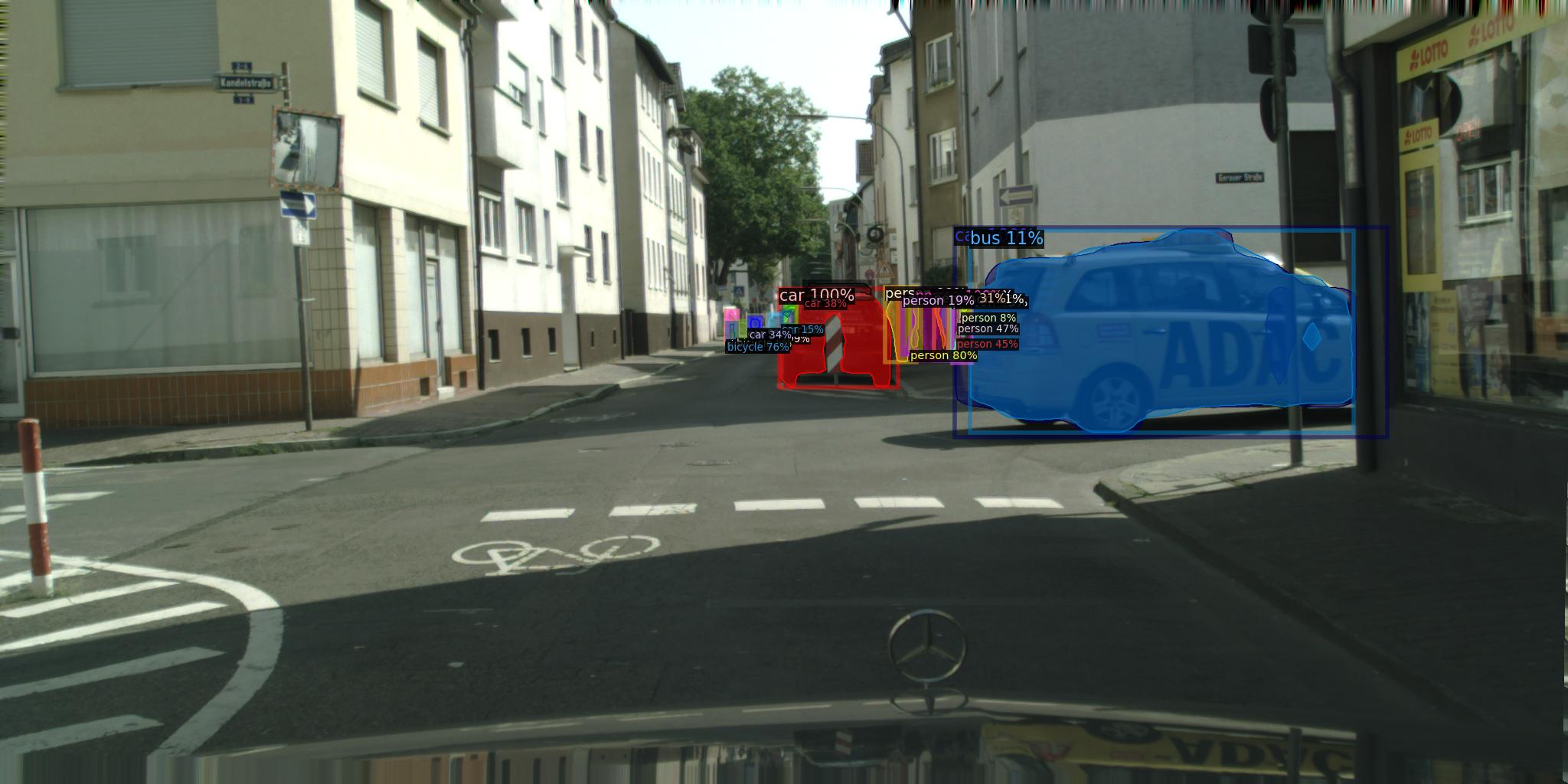} &
        \includegraphics[width=\myqwidth]{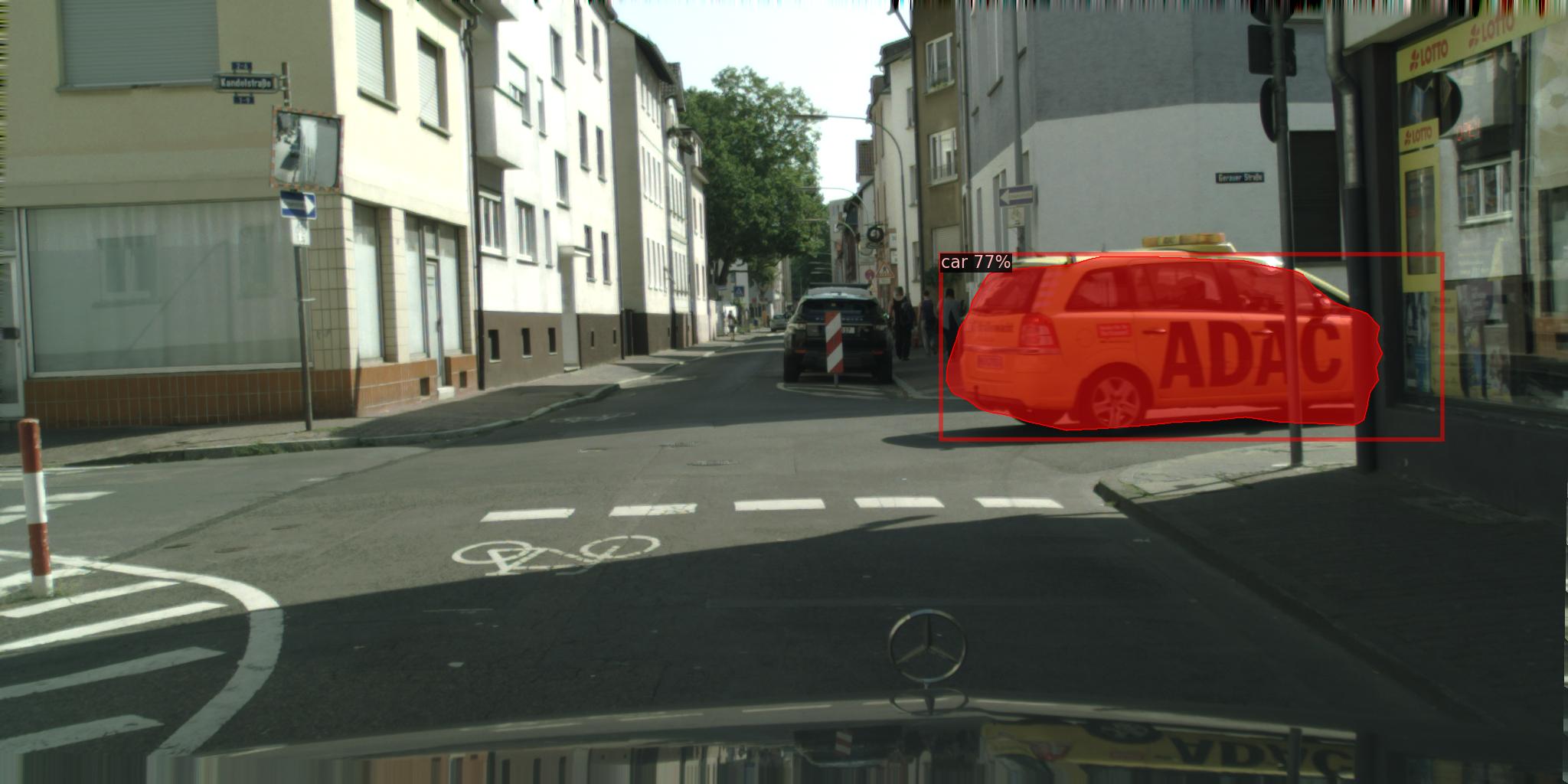} &
        \includegraphics[width=\myqwidth]{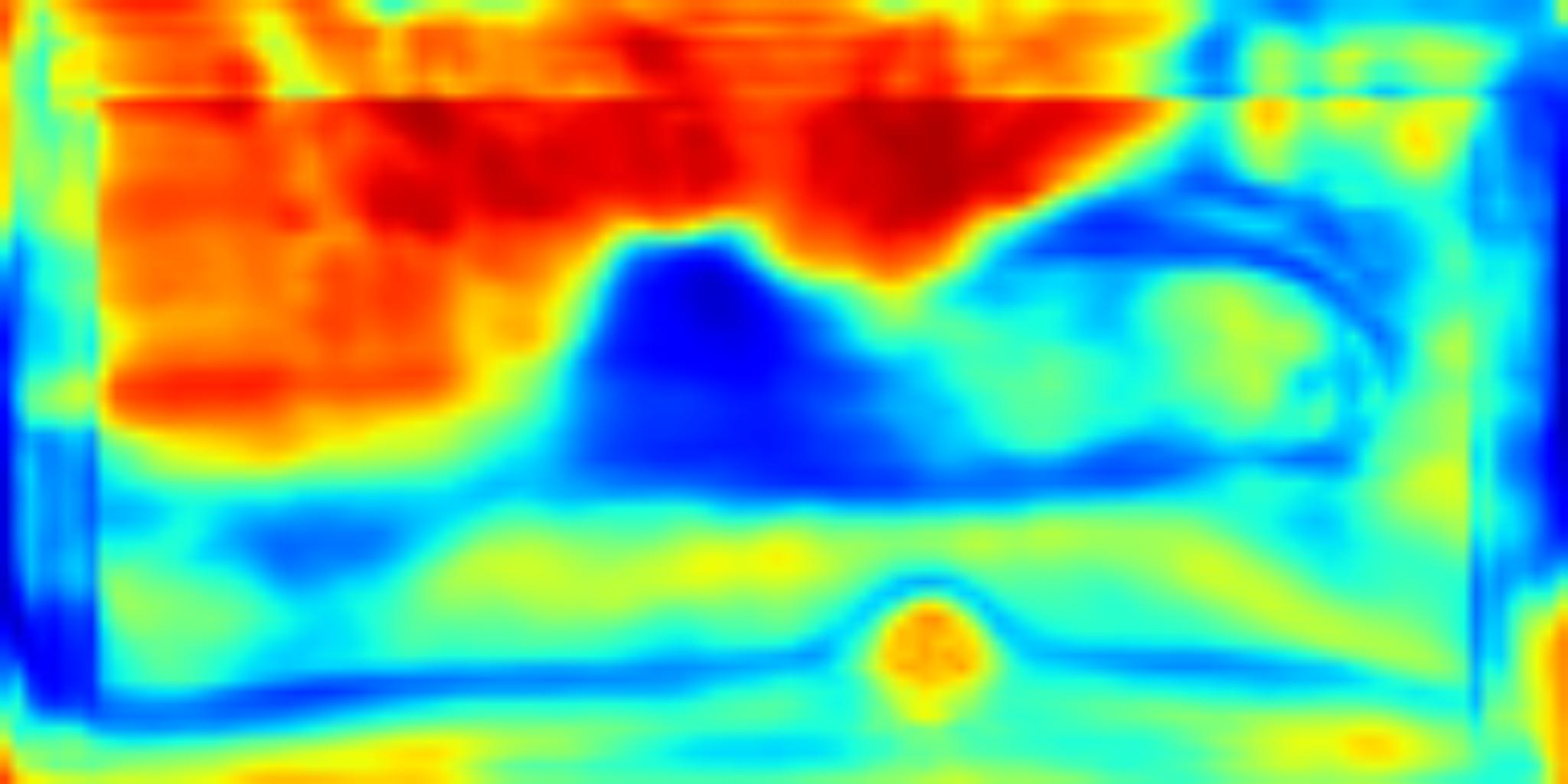} \\ [0.2em]
        
        \includegraphics[width=\myqwidth]{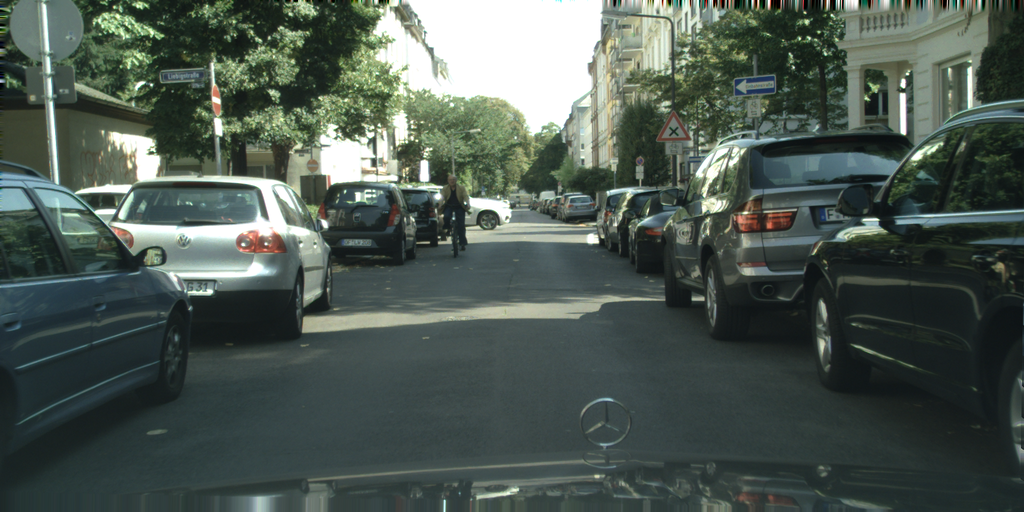} &
         \includegraphics[width=\myqwidth]{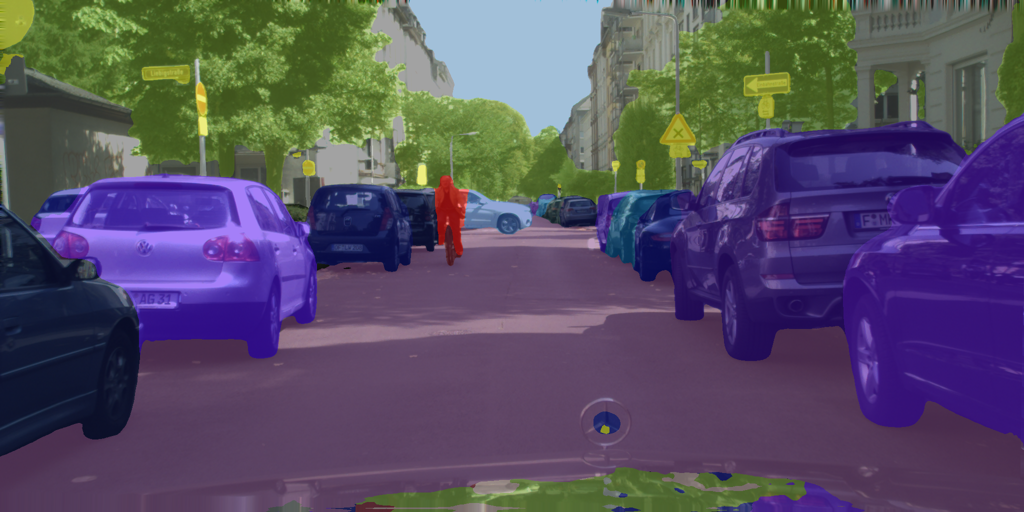} &
         \includegraphics[width=\myqwidth]{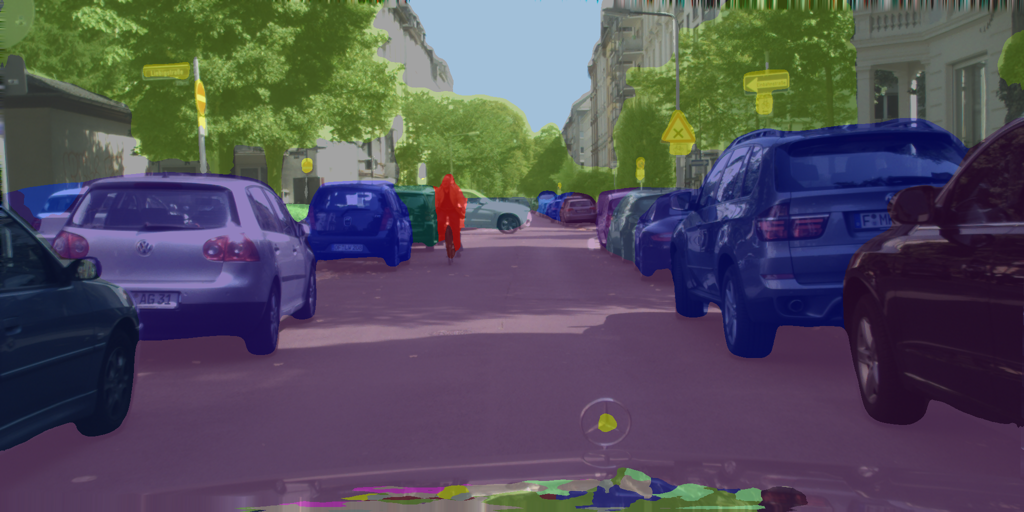} &
         \includegraphics[width=\myqwidth]{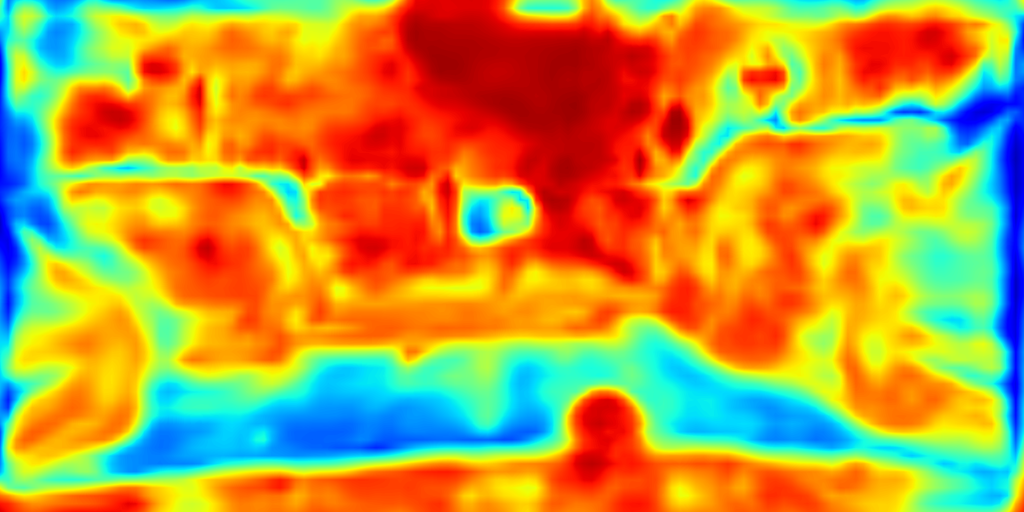} \\ [-0.1em]
         \includegraphics[width=\myqwidth]{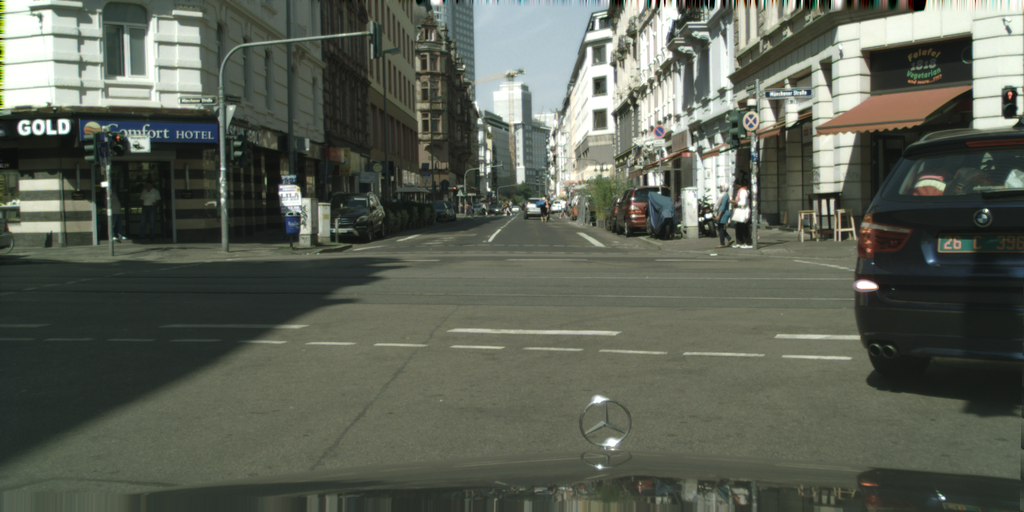} &
         \includegraphics[width=\myqwidth]{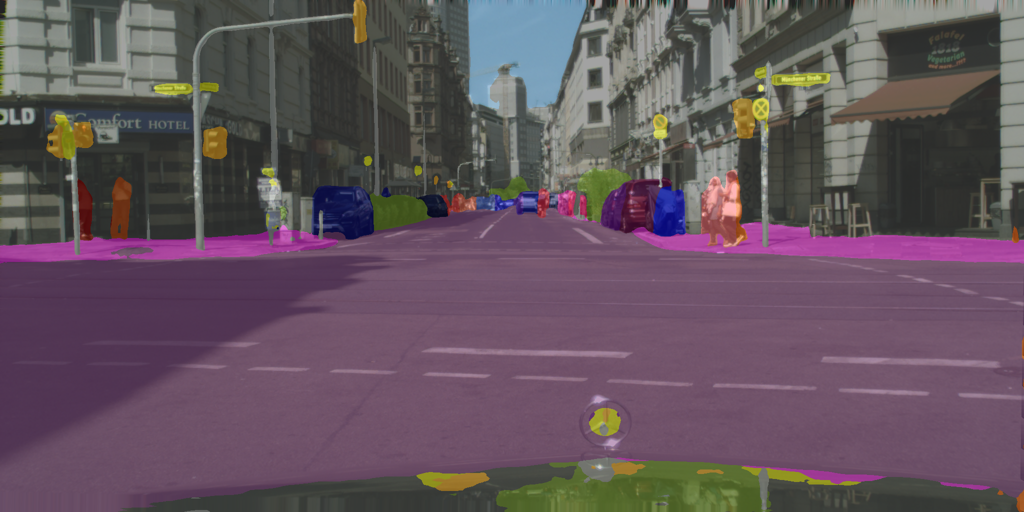} &
         \includegraphics[width=\myqwidth]{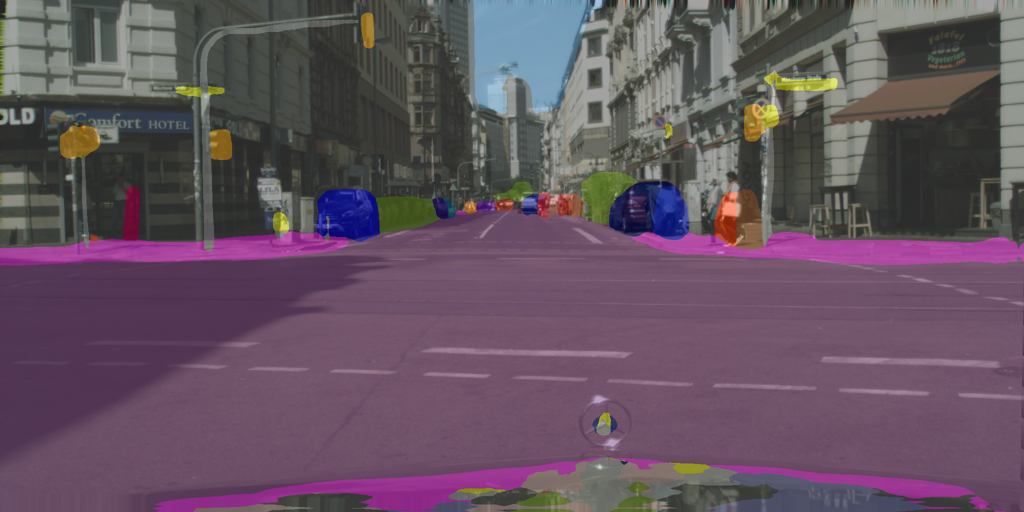} &
         \includegraphics[width=\myqwidth]{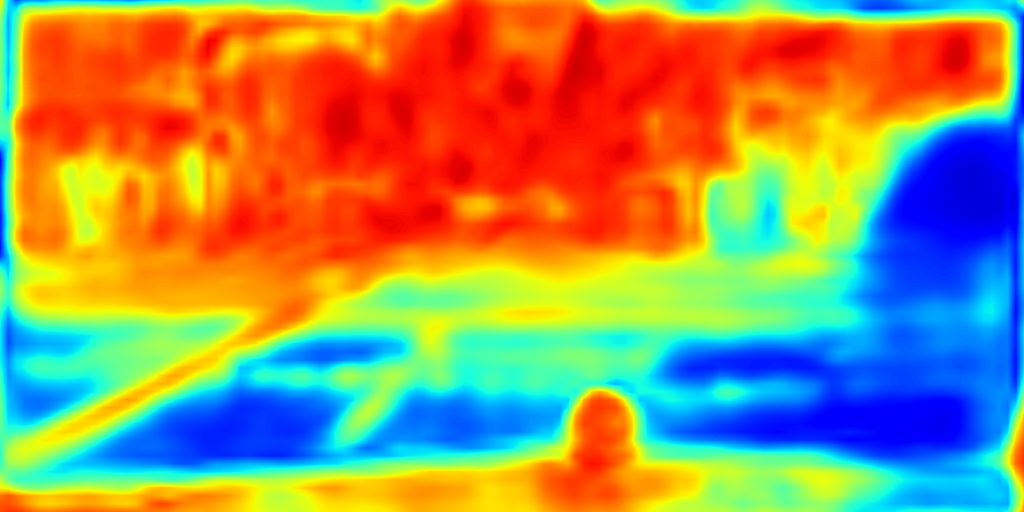}
    \end{tabular}
    \caption{
        Qualitative results 
        of dense semantic forecasting:
        semantic segmentation (top two rows),
        instance segmentation (middle two rows) and
        panoptic segmentation (bottom two rows). 
        We show one short-term and one
        mid-term example for each modality. 
        The columns show the last observed frame, 
        oracle prediction in the future frame, 
        F2MF forecast and the F2M heatmap.
    }       
    \label{fig:dense_pred}
\end{figure*}
Figure \ref{fig:dense_pred} visualizes 
our short-term and mid-term forecasts
and compares them with 
our oracle.
The columns show the last observed frame, 
the oracle prediction overlayed 
on top of the future frame,
our F2MF forecast overlayed 
on top of the future frame,
and the F2M heatmap 
$\beta^\textrm{F2M}$ = 
$1-\beta^\textrm{F2F}$ =
$\sum_\tau \beta^\textrm{F2M}_\tau$.
The red regions correspond to pixels forecasted 
by warping with the F2M flow 
while the blue regions correspond 
to pixels forecasted by the F2F module.
The blue pixels usually correspond 
to unoccluded scenery which 
has to be imagined by the model
since it was not visible
in any of the observed frames.
We observe that 
the forecasted F2M heatmaps are quite accurate 
in all images, which suggests that F2MF 
is versatile and task agnostic.

The top two rows of Figure~\ref{fig:dense_pred} 
correspond to semantic segmentation.
Most pixels from row 1 of this group 
are predicted by F2M forecasting
because there is little motion
in the scene.
On the other hand,
there is a large blue region
in the bottom left of 
the F2M heatmap in row 2.
The blue region was disoccluded by 
the car passing by the camera.
As it was never observed before,
the F2M module does not stand a chance,
so the F2F module has to in-paint
novel content.
We observe that the prediction is sound,
although it misses 
some of the people in behind.
We also provide a video presentation of our mid-term
forecasting performance on Frankfurt 
video\footnote{\url{https://www.youtube.com/watch?v=Wgo_5xnvn8M}}.

The middle two rows
illustrate instance forecasting
on Cityscapes val.
We observe that our
model accurately predicts
the future leg stance  
for some of the pedestrians in row 3.
The F2M heatmap reveals that most human pixels 
are forecasted by the F2F module 
except for the heads 
which exhibit more regular motion 
than the bodies.
Furthermore,
our model correctly
predicts the future position
of the moving taxi
in row 4.
The corresponding F2M heatmap
reveals that the F2F module
in-paints the features
in the disoccluded area.
Of course, we are unable to forecast
the objects behind the taxi.

The bottom two rows
address panoptic forecasting 
on Cityscapes val.
Notice that pixels corresponding to
object classes are colored in different shades
of the original class color.
We observe that the forecasted
segmentation appears sharp and accurate.
Similarly as in the semantic segmentation forecasting,
we observe that the F2F module
is in charge for novel regions.
This is best seen in the last row,
where the car on the right 
leaves the scene and dis-occludes
a large part of the background.
The model correctly forecasts
that the pixels behind the car
correspond to buildings,
sidewalk and road.

\subsection{Ablation and validation study}
Table~\ref{tab:main_ablation} shows
validation experiments which 
quantify contributions from 
our F2F, F2M and correlation modules.
We perform this study 
on semantic segmentation forecasting
with our single-frame model based on ResNet-18.
Note that we do not use 
data augmentation 
in this experiment.
This enables caching the features
on the SSD drive and therefore allows
significantly faster training.
Every model configuration presented 
in the table is separately trained.

First we compare F2F and F2M 
forecasting individually.
In experiments
without the correlation module,
F2F outperforms F2M 
for 0.6 mIoU points.
In experiments with the correlation module,
F2F outperforms F2M
for 0.7 mIoU points at short-term,
and they perform equally 
in mid-term forecast.
These results are plausible,
because the F2F approach
is more expressive.
While F2M has 
to explain the future
by warping past features,
F2F is unconstrained
and can imagine anything.
Nevertheless, 
the compound F2MF model
always outperforms any of the two
single-head approaches.
In a setup without the correlation module,
F2MF outperforms F2M by 1.0-1.2 mIoU points,
and the individual F2F by 0.4-0.6 mIoU points.
In a setup with the correlation module,
F2MF outperforms the F2M by 1.3-1.4 mIoU points,
and the  F2F by 0.6-1.4 mIoU points.
These results hint that 
F2M and F2F are indeed complementary.

The correlation module improves the performance
in all setups.
It contributes
0.8 and 2.3 mIoU points 
in short-term and mid-term F2M forecast respectively.
Likewise, the correlation module improves 
the F2F performance for 0.9 and 1.7 mIoU points
and boosts the compound F2MF model for 1.1-2.5 mIoU points 
at short-term and mid-term, respectively.

\begin{table}[h]
    \caption{
        Ablation of correlation, 
        F2F, and F2M on Cityscapes val.
        Standalone F2F and F2M models
        are trained independently.
    }
    \begin{center}
        
    \begin{tabular}{ccc|cc|cc}
      \toprule
         \multicolumn{3}{c|}{Configuration
          (F2MF-RN18)} & 
         \multicolumn{2}{c|}{Short-term mIoU} & \multicolumn{2}{c}{Mid-term mIoU} \\
         F2F    & F2M    & Correlation  & All & MO & All & MO \\
      \midrule
        & \cmark &        &         
        64.8  & 63.4 & 52.2 & 47.6 
      \\
        \cmark &        &        &         
        65.4  & 64.0 & 52.8 & 48.6 
      \\
        \cmark & \cmark &        &         
        65.8  & 64.7 & 53.4 & 49.7 
        \\
        & \cmark & \cmark &         
        65.6  & 64.4 & 54.5 & 50.7  
      \\
        \cmark &        & \cmark &         
        66.3  & 64.9 & 54.5 & 50.8  
      \\
        \cmark & \cmark & \cmark & 
        \textbf{66.9} & \textbf{65.6} &
        \textbf{55.9} & \textbf{52.4} 
      \\
      \bottomrule
    \end{tabular}
    \end{center}
    \label{tab:main_ablation}
\end{table}

\begin{figure}[h!]
    \begin{center}
    \begin{tabular}{@{}c@{}c@{}}
    \includegraphics[width=0.5\columnwidth]
      {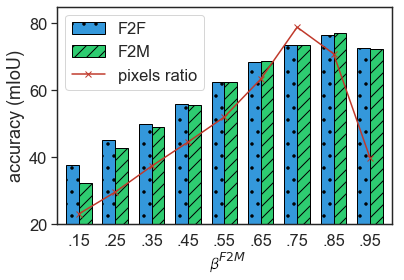} &
    \includegraphics[width=0.5\columnwidth]
      {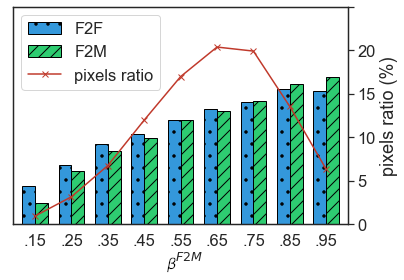} 
    \end{tabular}
    \end{center}
    \caption{
    Histograms of F2F and F2M accuracy and overall pixel incidence over $\beta^\textrm{F2M}$ bins as inferred by the compound F2MF model on Cityscapes val for short-term (left) and mid-term (right) forecasting.
    }
    \label{fig:f2mvsf2f}
\end{figure}

We further investigate the complementary nature
of the F2F and F2M approaches
by comparing the accuracy of 
independent single-head models
in stratified groups of pixels.
Previous results showed that the F2F model 
performs slightly better in general.
However, we know that predicting the future
in novel regions 
is particularly hard
for the F2M model.
Therefore, we hypothesise that 
F2M might perform better 
in the previously observed scenery.
We divide pixels into bins according 
to the $\beta^\textrm{F2M}$ weights
as predicted by the compound F2MF model,
and then show the forecasting accuracy
of the two independently trained
F2F and F2M models in Fig.~\ref{fig:f2mvsf2f}.
On the left y-axis we show
per-bin forecasting accuracy.
On the right y-axis we show
the relative share of pixels 
in that particular bin.
We show the F2M weights $\beta^\textrm{F2M}$
on the x-axis for each bin.
The figure separately considers 
the short-term (left) 
and mid-term (right) forecasting.  
We observe that the pixel histogram is skewed 
towards higher $\beta^\textrm{F2M}$ values. 
This suggests that the F2MF model 
delegates the majority of the pixels 
to the F2M module.
We also observe that this effect
is less pronounced in mid-term forecast.
This makes sense because
we observe more novelty there.
The plot also shows that F2M model performs better
in pixels with higher $\beta^\textrm{F2M}$ values,
which justifies the delegating decision 
from the compound model.
These plots support our hypothesis
that the two approaches are complementary.

Table~\ref{tab:numframes_abl} validates
the number of input frames
by exploring F2MF accuracy 
for different look-back windows.
Models which observe only 
the most recent frame
perform poorly as 
they are unable to
estimate the scene dynamics
with a single reference point.
Note that these models
do not have the correlation module
because at least two frames
are needed to compute
spatio-temporal coefficients.
Models with 2 to 5 input frames
perform comparably well,
while the model with 
4 input frames
performs the best.
This confirms the suitability
of the default forecasting setup
\cite{luc2017predicting}
on Cityscapes.
Please note that mid-term experiments
with 5 input frames
are not feasible for $\Delta$t=3
due to limited length of Cityscapes clips.

\begin{table}[h]
    \caption{
        Validation of the number 
        of input frames on Cityscapes.
    }
    \begin{center}
        
    \begin{tabular}{l|cc|cc}
      \toprule
         \multicolumn{1}{c|}{Input frames} & 
         \multicolumn{2}{c|}{Short-term mIoU} & \multicolumn{2}{c}{Mid-term mIoU} \\
         \multicolumn{1}{c|}{(F2MF-RN18)}
          & All & MO & All & MO \\
      \midrule        
        $\{t\}$ & 57.9 & 55.5 & 45.7 & 39.6 \\
        $\{t-3, t\}$ & 66.4 & 65.3 & 54.9 & 51.2 \\
        $\{t-6, t-3, t\}$ & 66.9 & 65.6 & 55.1 & 51.0 \\
        $\{t-9, t-6, t-3, t\}$ & \textbf{66.9} & \textbf{65.6} & \textbf{55.9} & \textbf{52.4} \\
        $\{t-12, t-9, t-6, t-3, t\}$ & 66.7 & 65.1 & / & / \\
      \bottomrule
    \end{tabular}
    \end{center}
    \label{tab:numframes_abl}
\end{table}

We have also attempted to forecast 
3 and 9 steps into the future
by observing input frames
with stride 2
($\tau \in \{t-6, t-4, t-2, t\}$).
Our F2MF-RN18 model achieved
$67.3$ mIoU in short-term forecast
and $55.6$ mIoU at mid-term.
Thus, forecasting with input stride 2
improved for $0.4$ points 
with respect to the standard setup
at short-term,
while underperforming  
for $0.3$ points at mid-term.
This suggests that mid-term forecasting 
profits from a larger look-back window, 
which can also be observed in Table \ref{tab:numframes_abl}.

Table \ref{tab:weight_abl} validates
the contribution of 
the proposed weight module
which outputs 
five dense feature maps
used to blend 
the four F2M forecasts
and the F2F forecast.
We compare the proposed design
with two baselines.
The first baseline simply averages 
all of the five feature forecasts.
The second baseline predicts
five scalars which 
represent the weights 
for each of the five 
future feature predictions.
This baseline uses tensor-wide
instead of the original per-pixel blending.
The table shows that tensor-wide weights
outperform simple averaging 
by 1.4 mIoU points in short-term
and 0.7-1.3 mIoU points 
in mid-term forecast.
The proposed per-pixel weights
further improve the performance
for 0.2-0.5 mIoU points in short-term
and 0.6-0.7 mIoU points in mid-term forecast.

\begin{table}[h]
    \caption{
        Validation of the blending weights module on Cityscapes val.
    }
    \begin{center}
    \begin{tabular}{lcccc}
      \toprule
        Blending method & \multicolumn{2}{c}{Short-term mIoU} & \multicolumn{2}{c}{Mid-term mIoU} \\
        \multicolumn{1}{c}{(F2MF-RN18)} & All & MO & All & MO \\
        \midrule
        mean weights & 65.3 & 63.7 & 54.6 & 50.4 \\
        image-level weights & 66.7 & 65.1 & 55.3 & 51.7 \\
        per-pixel weights & \textbf{66.9} & \textbf{65.6} & \textbf{55.9} & \textbf{52.4} \\
      \bottomrule
    \end{tabular}
    \end{center}
    \label{tab:weight_abl}
\end{table}
\subsection{Autoregressive long-term forecasting}

\begin{figure}[h]
    \centering
    \includegraphics[width=\columnwidth]{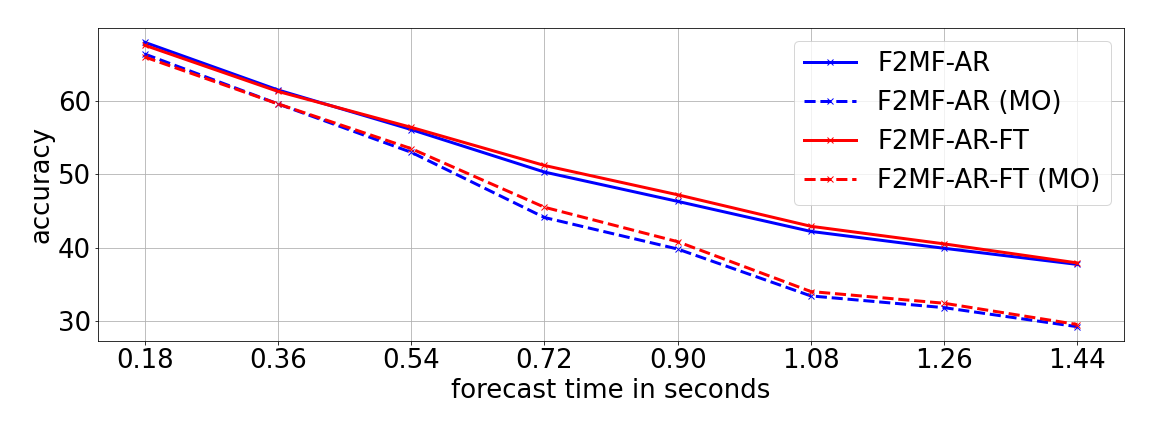}
    \caption{
        Dependence of the semantic segmentation accuracy 
        of our autoregressive models
        on different forecasting offsets
        on Frankfurt videos from Cityscapes val.
        \emph{FT} denotes autoregressive fine-tuning.
        }
    \label{fig:longterm}
\end{figure}

We investigate autoregressive application
of our short-term model for long-term forecasting.
Each subsequent invocation of our forecasting approach 
takes the precedent forecast as the most recent input.
This enables forecasting arbitrary 
number of timesteps into the future.
We evaluate the performance
only on Frankfurt scenes from Cityscapes val
because video clips 
from other cities have only very short videos.
We fine-tune our best short-term model 
for autoregressive forecast
by accumulating the loss in timesteps
$t+3$, $t+6$ and $t+9$,
and backpropagating gradients through time.
Figure~\ref{fig:longterm} shows 
the mIoU accuracy
for different forecasting times
both for the regular (F2MF-AR)
and the fine-tuned model (F2MF-AR-FT).

\subsection{Cross-dataset generalization}
We investigate
whether our feature-based forecasting 
generalizes to unseen data
beyond the validation subset,
as proposed in \cite{luc2017predicting}.
We aim to test our models on the CamVid 
dataset \cite{brostow2009semantic}
which differs 
from Cityscapes in terms of 
image resolution, 
video framerate and 
the set of semantic labels.
No adaptation is needed 
regarding the image resolution 
because our model is fully convolutional.
We sample CamVid videos each five timesteps,
which is approximately equal to 
three timesteps in Cityscapes.
We evaluate the mIoU accuracy 
according to a mapping 
from 19 Cityscapes classes 
to 11 CamVid classes.

Table~\ref{tab:camvid} shows the accuracy
of our single-step and autoregressive models,
and also the autoregressive approach from \cite{luc2017predicting}.
The columns show the oracle accuracy (mIoU), 
the forecast accuracy (mIoU) as well as the 
relative performance drop
w.r.t.\ to the oracle model.
We observe that our approach
achieves the highest accuracy
and also the lowest relative performance drop.
\begin{table}[h]
    \caption{
        Semantic segmentation forecasting on CamVid dataset with models trained on Cityscapes.
    }
    \begin{center}
    \begin{tabular}{lcccc}
        \toprule
        & Oracle & Forecast  & Rel.\ perf. & Drop  \\
        \midrule
        Luc ar.\ ft. 
        \cite{luc2017predicting} 
        & 55.4 & 46.8 & 84.5\% & -15.5\% \\
        \midrule
        F2MF-DN121 
        & 62.8  & 51.3 & 81.7\% & -18.3\% \\
        F2MF-DN121 ar. 
        & 62.8 & 53.4 & 85.0\% & -15.0\% \\
        F2MF-DN121 ar.\ ft. 
        & 62.8 & \textbf{54.5} & \textbf{86.8\%} & \textbf{-13.2\%} \\
        \bottomrule
    \end{tabular}
    \end{center}
    \label{tab:camvid}
\end{table}

Figure~\ref{fig:camvid} shows the performance 
of our best model from Table \ref{tab:camvid} on two CamVid scenes.
The rows show the last raw frame,
ground truth and 
the forecasted segmentation.
We observe that the feature forecasting succeeds 
to overcome the domain shift to a considerable degree.

\newcommand{\mycwidth}{0.33\columnwidth}
\begin{figure}[h]
    \centering
        \begin{tabular}{@{}c@{\,}c@{\,}c@{}}
         \includegraphics[width=\mycwidth]{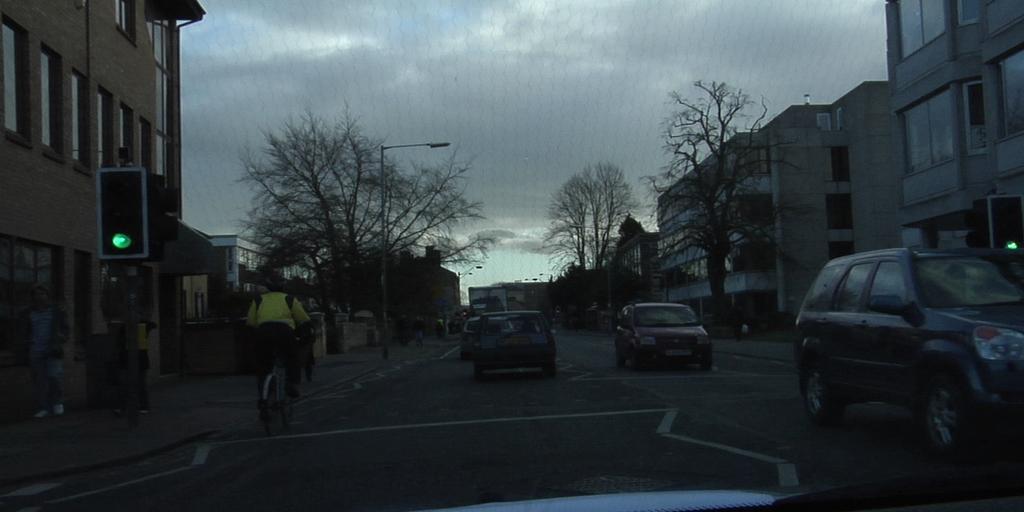} &
         \includegraphics[width=\mycwidth]{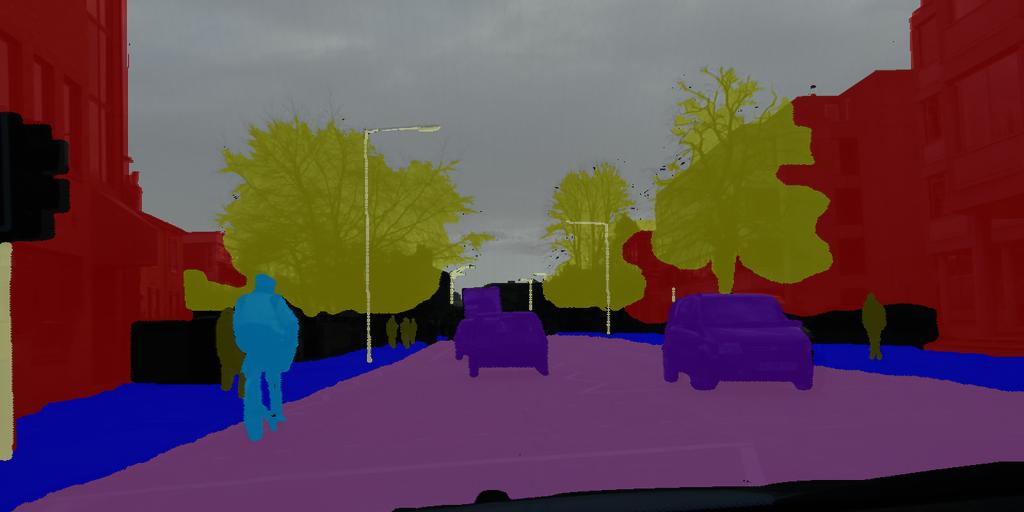} &
         \includegraphics[width=\mycwidth]{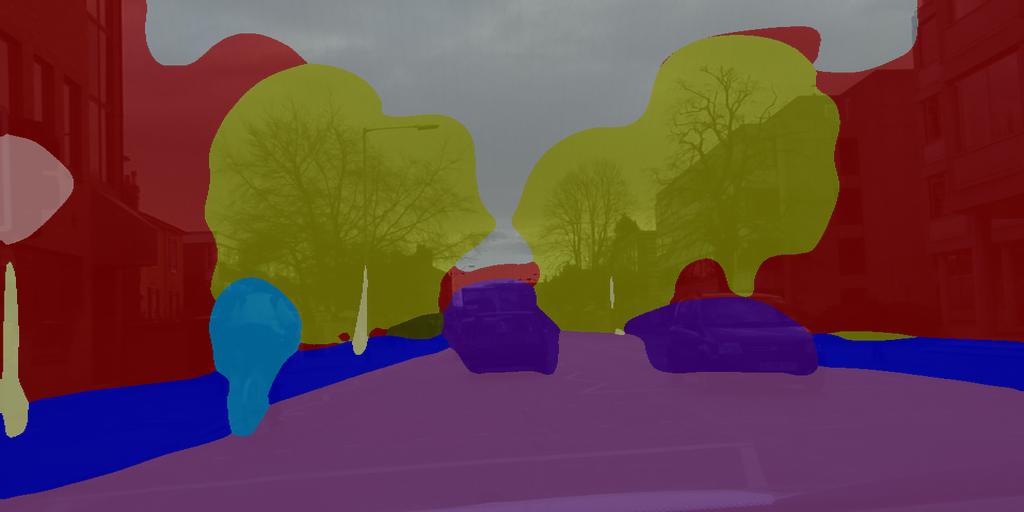} \\ [-0.25em]
         \includegraphics[width=\mycwidth]{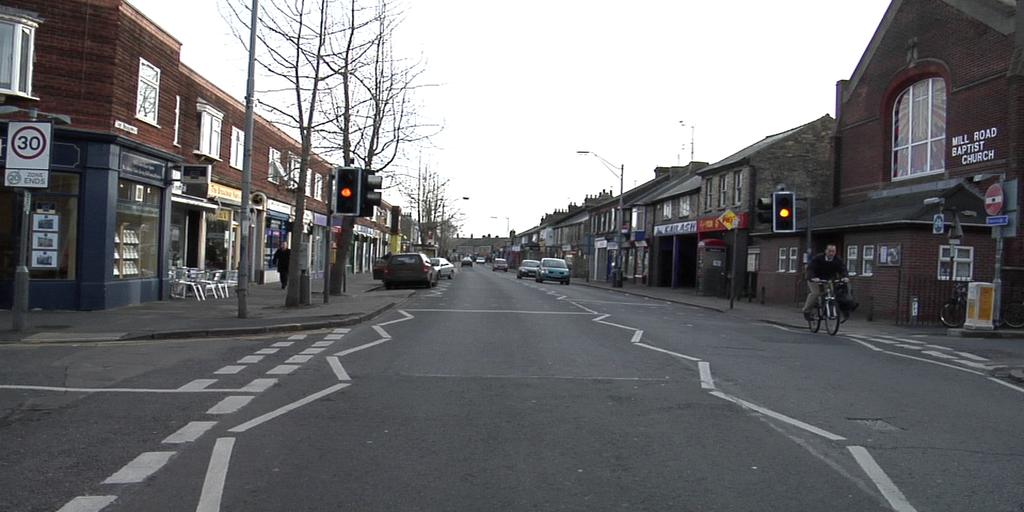}  & 
         \includegraphics[width=\mycwidth]{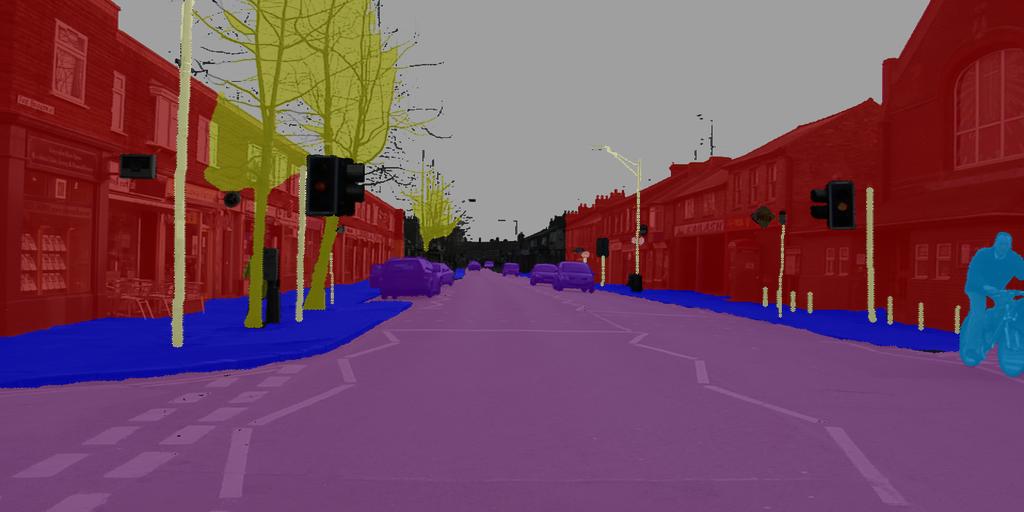}  &
         \includegraphics[width=\mycwidth]{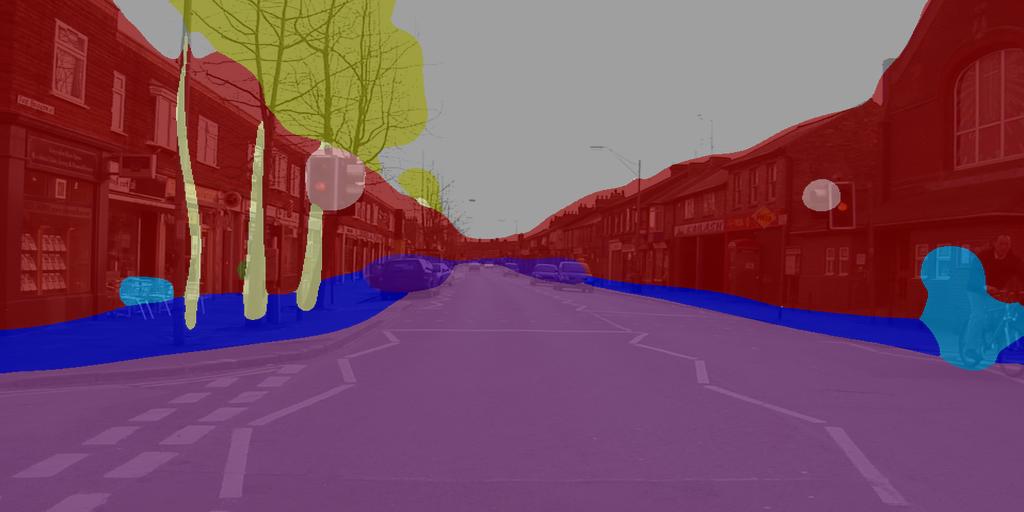} \\
    \end{tabular}
    \caption{Mid-term forecast on two scenes 
        from the CamVid dataset as produced 
        by our best model from Table \ref{tab:camvid}. 
        The columns show the last observed frame, 
        future ground truth and our F2MF forecast.}
    \label{fig:camvid}
\end{figure}

\subsection{Interpreting the model decisions}
\begin{figure*}[h!]
    \centering
    \begin{tabular}{@{}c@{}}
    \includegraphics[width=.999\textwidth]{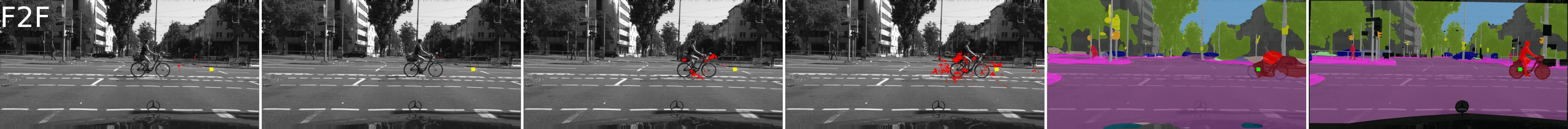}
    \\[-0.2em]
    \includegraphics[width=.999\textwidth]{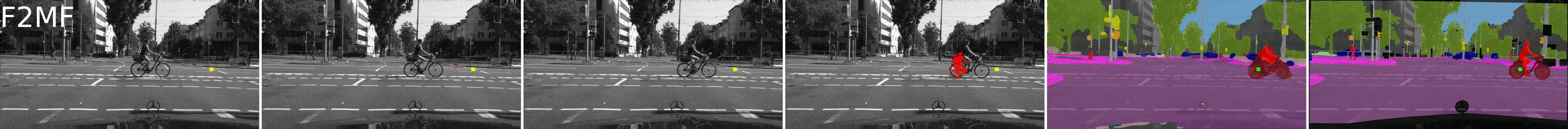}\\
    \\[-1.5em]
    \includegraphics[width=.999\textwidth]{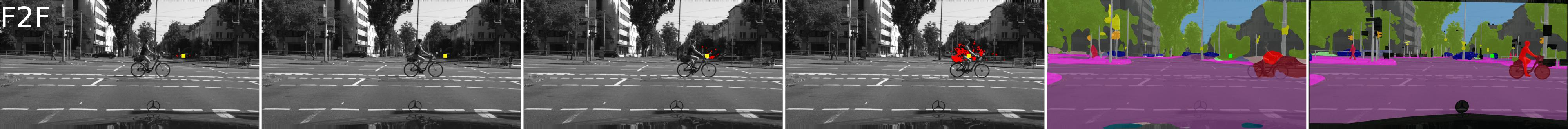}
    \\[-0.2em]
    \includegraphics[width=.999\textwidth]{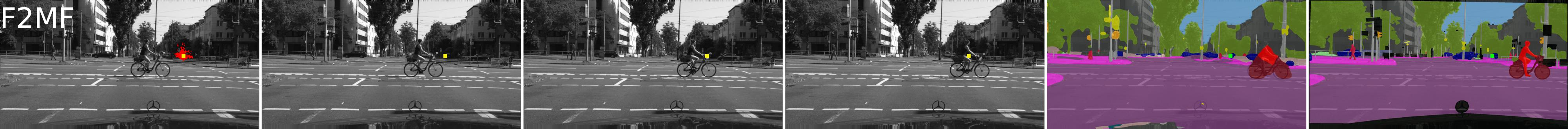}\\
    \end{tabular}

    \caption{
      Interpretation of F2F (rows 1,3)
      and F2MF (rows 2,4) decisions 
      in two pixels denoted
      with green squares.
      We consider a pixel on the 
      bicycle (rows 1-2)
      and a pixel on disoccluded 
      background (rows 3-4).  
      The columns show the four input frames,
      the forecasted semantic map,
      and the groundtruth
      with the overlayed future frame.
      Yellow squares show the
      positions of the considered two pixels
      in the observed frames.
      Red dots show top gradients 
      of the green pixel max-log-softmax
      w.r.t.\ input. These dots correspond to input pixels which are most responsible for the model decision in the green pixel.
      We observe that the F2MF forecast establishes a better spatio-temporal focus than the F2M forecast.
    }
    \label{fig:gradients}
\end{figure*}
This subsection further investigates the difference
between F2MF and F2F forecasting for semantic segmentation.
Figure~\ref{fig:gradients} considers two particular output pixels and visualizes 
pixels with top 1\textperthousand \ gradients
of log-max-softmax 
w.r.t.\ to the input frames (red dots).
Columns 1-4 show the four input frames.
Columns 5-6 show
our forecast
and the ground truth.
Pairs of rows correspond to two output pixels
which are designated with a green square in the predictions
and with a yellow square in the observed frames.
Rows 1-2 consider a pixel located 
at the rear bicycle wheel 
in the future frame.
We observe that 
most of the gradients are 
located at the corresponding spot 
in the most recent frame
for both models.
However,
the gradients
of the F2MF model
are less spread 
and more focused
towards the corresponding input pixel.
Rows 3-4 consider a pixel
which is occluded by the passing bicyclist in the most recent
frame, while being disoccluded in the future frame.
We observe that the F2F model
relies on the context
by looking around the pixel location
in the most recent frame.
On the other hand, 
the F2MF model realizes that
this part of the scene
had been visible
in the most distant frame.
Thus, our F2MF module makes the forecast
just by copying the representation
from the most distant feature tensor.
This shows that our 
F2MF model is able to learn 
to handle complex occlusion patterns. 

\subsection{Computational Complexity}
We analyze computational complexity 
of our methods 
and compare it with the corresponding 
state-of-the-art. 
We assume that the system is able
to cache the necessary past activations
and therefore consider the computational effort
which is required to make the forecast
after acquiring the current RGB frame.
We report the count of 
multiply-accumulate operations (MAC)
as measured with \texttt{thop} \cite{thopgithub}.

\begin{table}[b]
    \caption{Computational complexity of 
      semantic segmentation forecasting (GMAC)
      on full-resolution Cityscapes images.
    }
    \centering
    \begin{tabular}{lrr}
             \toprule
         Modules 
         & M2M \cite{terwilliger2019recurrent} 
         & F2MF-DN121 \\
         \midrule
        Single-frame semantics & 536.4 & 145.5 \\
        Flow (FlowNet2-C)      & 88.9  & 0.0 \\
        Forecasting            & 38.8  & 9.8 \\
        \midrule
        Total forecasting      & 127.7 & 9.8 \\
        Total                  & 664.1 & 155.3 \\
        \bottomrule
    \end{tabular}
    \label{tab:semseg_macs}
\end{table}
Table~\ref{tab:semseg_macs} compares
our F2MF-based semantic segmentation forecasting
with the runner-up method 
\cite{terwilliger2019recurrent}
from Table~\ref{tab:semseg}.
We observe that our forecasting incurs
$12\times$ less complexity.
The advantage remains considerable ($4\times$)
even if
we consider a comparison with
single-frame semantics included.
The difference mostly stems from the fact
that our forecasting model 
operates on 1/32 resolution
and does not require 
high resolution optical flow.
The approach \cite{terwilliger2019recurrent} 
has to evaluate PSP-Net 
and FlowNet2-C at 0.5 MPx,
as well as to forecast
flow features on 1/8 resolution.
Our estimate assumes that 
their forecast requires 
evaluation of a single ConvLSTM cell 
with 74 feature maps 
both in the hidden state 
and the input 
\cite{terwilliger2019recurrent}.

Table~\ref{tab:insseg_macs} compares
our F2MF-based instance-level forecasting
with the multi-level F2F approach \cite{luc2018predicting}.
The table shows the cost of single-frame inference
and the cost of forecasting at each subsampling level.
Our single-frame inference is a bit inefficient
and requires around $1.6\times$ more compute
than FPN based Mask R-CNN.
Nevertheless, our total forecasting cost is
$18\times$ smaller than
multi-level F2F \cite{luc2018predicting}.
This is because we only forecast features
which are $16\times$ subsampled
w.r.t.\ to the input resolution.
On the other hand,
multi-level approaches
\cite{luc2018predicting, sun2019predicting}
additionally have to forecast
expensive fine-resolution features at
1/8$\times$ and 1/4$\times$ resolution.
We expect that \cite{sun2019predicting}
would suffer from the same problem
although their source code is not available.
\begin{table}[h]
    \caption{Computational complexity of 
      instance segmentation forecasting (GMAC).}
    \centering
    \begin{tabular}{lrr}
        \toprule
        Modules & F2F-M-FPN-RN50 \cite{luc2018predicting} & F2MF-M-C4-RN50 \\
        \midrule
        Single-frame sem. & 401.0  & 668.4 \\
        Forecasting at 1/4$\times$ & 1417.7 & 0.0 \\
        Forecasting at 1/8$\times$ & 354.4  & 0.0 \\
        Forecasting at 1/16$\times$ & 88.6   & 106.2 \\
        Forecasting at 1/32$\times$ & 22.1   & 0.0 \\
        \midrule
        Total Forecasting & 1865.2 & 106.2 \\
        Total & 2266.2 & 774.6 \\
        \bottomrule
    \end{tabular}
    \label{tab:insseg_macs}
\end{table}

Table~\ref{tab:execution_profile} 
shows unoptimized execution profiles 
of our forecasting models
on a GTX1080Ti GPU.
We measure the average time
for three inference stages:
feature extraction in four input frames,
feature forecasting, 
and semantic formation.
Sections correspond to three
dense prediction tasks:
semantic segmentation (top),
instance segmentation (middle),
and panoptic segmentation (bottom).
We observe that forecasting
is much faster than feature extraction 
and single-frame inference.
F2MF-RN18 allows
real-time inference
since online execution environments 
allow to cache the past features
and to perform feature extraction 
in only one frame --- the most recent one.

\begin{table}[h]
  \caption{Execution profile of 
    unoptimized F2MF models
    (milliseconds).}
  \centering
  \begin{tabular}{lccc}
    \toprule
    Model & 4$\times$ Extraction & 
            Forecasting & Semantics \\
    \midrule
      F2MF-RN18  & 72 & 7 & 7 \\
      F2MF-DN121 & 265 & 12 & 8 \\ 
      \midrule
      F2MF-Mask-C4-RN50 & 204 & 47 & 248 \\
      \midrule
      F2MF-PDL-RN50 & 230 & 52 & 144 \\
      \bottomrule
  \end{tabular}
  \label{tab:execution_profile}
\end{table}
\section{Conclusion}

Anticipation of future semantics is a prerequisite
for intelligent planning of current actions.
Recent work addresses this problem by
implicitly capturing laws of scene dynamics
throughout deep learning in video.
However, the existing approaches
are unable to distinguish
disoccluded and emerging scenery
from previously observed parts of the scene.
This is suboptimal,
since the former requires pure recognition
while the latter can be explained by warping.
Different than all previous approaches,
our method predicts
emergence of unobserved scenery
and exploits that information
for disentangling variation caused by novelty
from variation due to motion.

Our method performs dense semantic forecasting
on the feature level.
Different than previous such approaches,
we regularize the forecasting process
by expressing it as a causal relationship
between the past and the future.
The proposed F2M (feature-to-motion) forecasting
generalizes better than the classic
F2F (feature-to-feature) approach
at many (but not all) image locations.
We achieve the best of both worlds
by blending F2M and F2F predictions
with densely regressed weight factors.
We empirically confirm that low F2M weights
occur at unobserved scenery.
The resulting F2MF approach
surpasses the state-of-the-art
in semantic segmentation forecasting
on the Cityscapes dataset by a wide margin.

We complement convolutional features
with their respective correlation coefficients
organized within a cost volume
over a small set of discrete displacements.
Our forecasting models use deformable convolutions
in order to account for geometric nature of F2F forecasting.
These two improvements bring clear advantage
in all three feature-level approaches: F2F, F2M, and F2MF.
To the best of our knowledge, this is the first account
of using these improvements for semantic forecasting.

Unlike previous methods,
our single-frame model for semantic segmentation
does not use skip connections along the upsampling path.
Consequently, we are able to forecast
condensed abstract features
at the far end of the downsampling path
with a single F2MF module.
This greatly improves the inference speed
and favors the forecasting accuracy
due to coarse resolution and
high semantic content of the involved features.
We were unable to outperform this approach
with multi-level F2F forecasting
in spite of significantly better
single-frame accuracy.

We also propose an adaptation of the proposed F2MF method
for two additional dense prediction tasks:
instance segmentation and panoptic segmentation.
These experiments use third-party
single-frame models
and therefore show that
our method can be successfully used
as a drop-in solution for converting
any kind of dense prediction model
into its competitive forecasting counterpart.
To the best of our knowledge, this is the first account of panoptic forecasting in the scientific literature.

The proposed method offers many exciting directions for future work.
In particular, our method does not address multi-modal future,
which is a key to long-term forecasting and worst-case reasoning.
Other suitable extensions include overcoming obstacles
towards end-to-end training, extension to RGB forecasting, 
as well as enforcing temporally consistent predictions 
in neighbouring video frames.

\bibliographystyle{IEEEtran}
\bibliography{egbib.bib}

\begin{IEEEbiography}[{\includegraphics[width=1in,height=1.25in,clip,keepaspectratio]{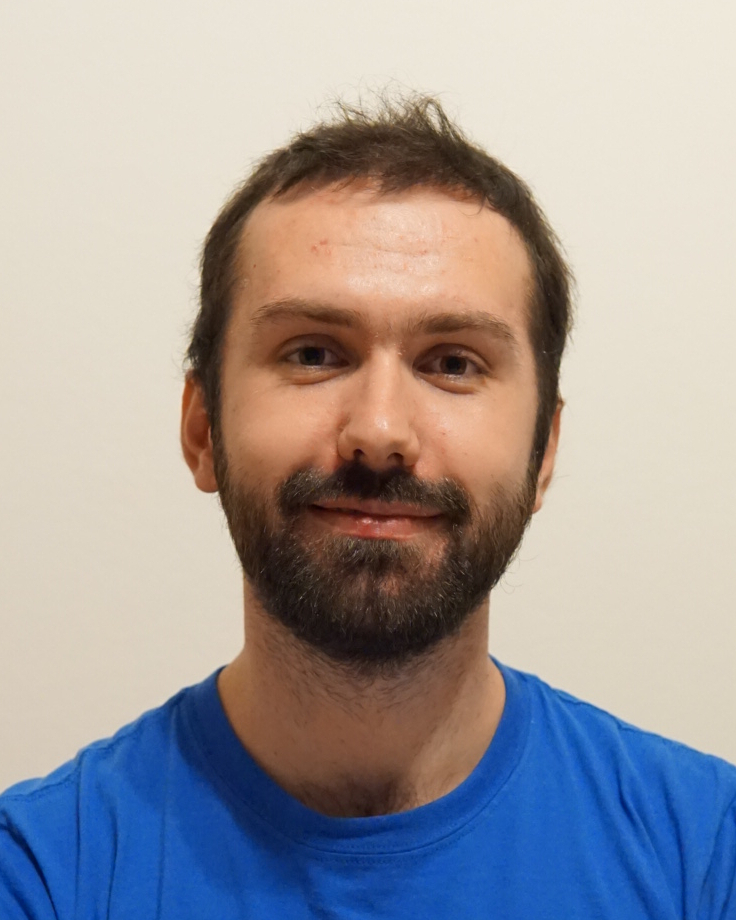}}]{Josip Šarić}
received the M.Sc. degree in computer
science from the University of Zagreb, Croatia. 
He is currently a Research Assistant
at the Faculty of Electrical Engineering and
Computing, University of Zagreb. 
His research interests include 
efficient convolutional architectures 
for image recognition
and dense semantic forecasting.
\end{IEEEbiography}

\begin{IEEEbiography}[{\includegraphics[width=1in,height=1.25in,clip,keepaspectratio]{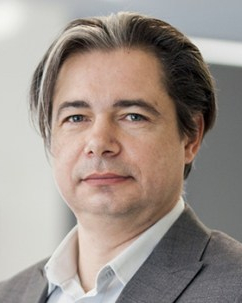}}]{Sacha Vražić}
is the Director of Autonomous Driving R\&D at Rimac Automobili, 
where he works on solving issues 
to achieve the full autonomy 
for self-driving vehicles.
He was heading the research laboratory 
for a Toyota Group company, 
and before he was teaching 
at University in Nice, France. 
He authored number of publications and patents. 
His research interest are in computer vision,
machine learning and blind source separation.

\end{IEEEbiography}

\begin{IEEEbiography}[{\includegraphics[width=1in,height=1.25in,clip,keepaspectratio]{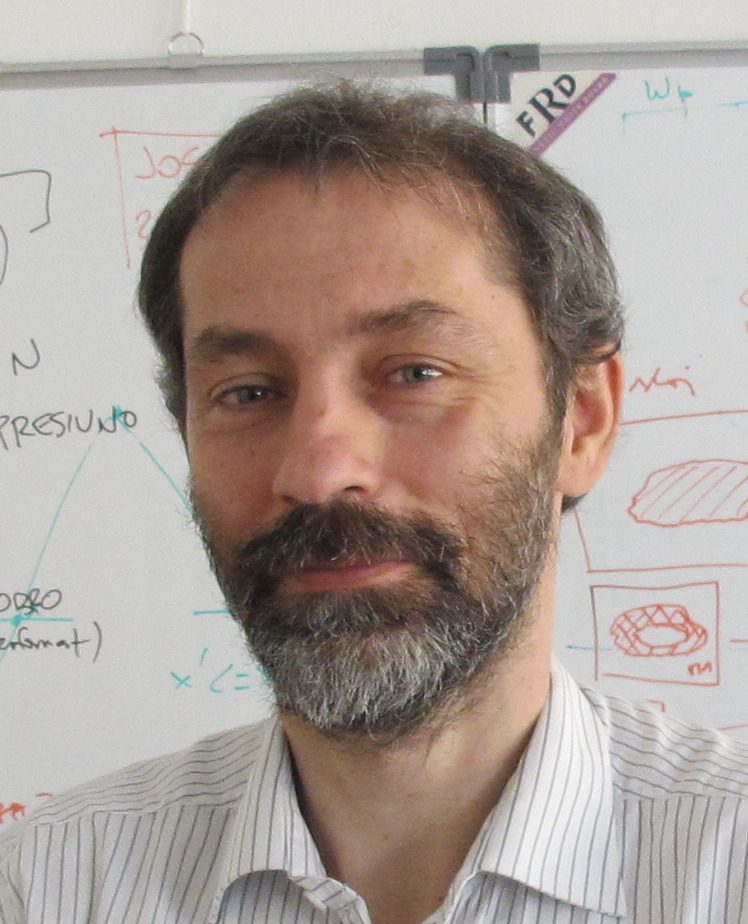}}]{Siniša Šegvić}
received the Ph.D. degree in computer
science from the University of Zagreb, Croatia. He
was a Post-Doctoral Researcher at IRISA Rennes,
for one year, and as a Post-Doctoral Researcher at
TU Graz, for one year. He is currently a Full Professor at the Faculty of Electrical Engineering and
Computing, University of Zagreb. His research interests include efficient convolutional architectures for
classification, dense prediction, and dense semantic
forecasting.
\end{IEEEbiography}

\end{document}